\documentclass[10pt,twocolumn,twoside]{IEEEtran}
\usepackage{cite}
\usepackage{amsmath,amssymb,amsfonts}
\usepackage{multirow}
\usepackage{graphicx}
\usepackage{textcomp}
\usepackage{xcolor}

\usepackage{subcaption}
\usepackage{cleveref}

\usepackage{booktabs}
\usepackage{threeparttable}

\usepackage[lined,boxed,commentsnumbered]{algorithm2e}
\usepackage{algpseudocode}
\usepackage{amsmath}
\usepackage{amsthm}
\usepackage{comment}

\def\BibTeX{{\rm B\kern-.05em{\sc i\kern-.025em b}\kern-.08em
    T\kern-.1667em\lower.7ex\hbox{E}\kern-.125emX}}
\begin{document}
\newtheorem{definition}{\it Definition}
\newtheorem{theorem}{\bf Theorem}
\newtheorem{lemma}{\it Lemma}
\newtheorem{corollary}{\it Corollary}
\newtheorem{remark}{\it Remark}
\newtheorem{example}{\it Example}
\newtheorem{case}{\bf Case Study}
\newtheorem{assumption}{\it Assumption}
\newtheorem{property}{\it Property}
\newtheorem{proposition}{\it Proposition}

\newcommand{\hP}[1]{{\boldsymbol h}_{{#1}{\bullet}}}
\newcommand{\hS}[1]{{\boldsymbol h}_{{\bullet}{#1}}}

\newcommand{\ba}{\boldsymbol{a}}
\newcommand{\baq}{\overline{q}}
\newcommand{\bA}{\boldsymbol{A}}
\newcommand{\bb}{\boldsymbol{b}}
\newcommand{\bB}{\boldsymbol{B}}
\newcommand{\bc}{\boldsymbol{c}}
\newcommand{\bcG}{\boldsymbol{\cal G}}
\newcommand{\bcO}{\boldsymbol{\cal O}}
\newcommand{\be}{\boldsymbol{e}}
\newcommand{\tbe}{\tilde{\be}}
\newcommand{\bh}{\boldsymbol{h}}
\newcommand{\bH}{\boldsymbol{H}}
\newcommand{\bl}{\boldsymbol{l}}
\newcommand{\bm}{\boldsymbol{m}}
\newcommand{\bn}{\boldsymbol{n}}
\newcommand{\bo}{\boldsymbol{o}}
\newcommand{\bO}{\boldsymbol{O}}
\newcommand{\bp}{\boldsymbol{p}}
\newcommand{\tbp}{\tilde{\bp}}
\newcommand{\bq}{\boldsymbol{q}}
\newcommand{\tbq}{\tilde{\bq}}
\newcommand{\br}{\boldsymbol{r}}
\newcommand{\tbr}{\tilde{\br}}
\newcommand{\bR}{\boldsymbol{R}}
\newcommand{\bs}{\boldsymbol{s}}
\newcommand{\bS}{\boldsymbol{S}}
\newcommand{\bT}{\boldsymbol{T}}
\newcommand{\bv}{\boldsymbol{v}}
\newcommand{\bw}{\boldsymbol{w}}

\newcommand{\balpha}{\boldsymbol{\alpha}}
\newcommand{\bbeta}{\boldsymbol{\beta}}
\newcommand{\bomega}{\boldsymbol{\omega}}
\newcommand{\bOmega}{\boldsymbol{\Omega}}
\newcommand{\bTheta}{\boldsymbol{\Theta}}
\newcommand{\bphi}{\boldsymbol{\phi}}
\newcommand{\btheta}{\boldsymbol{\theta}}
\newcommand{\bvarpi}{\boldsymbol{\varpi}}
\newcommand{\bpi}{\boldsymbol{\pi}}
\newcommand{\brho}{\boldsymbol{\rho}}
\newcommand{\bpsi}{\boldsymbol{\psi}}
\newcommand{\bxi}{\boldsymbol{\xi}}
\newcommand{\bx}{\boldsymbol{x}}
\newcommand{\by}{\boldsymbol{y}}

\newcommand{\cA}{{\cal A}}
\newcommand{\bcA}{\boldsymbol{\cal A}}
\newcommand{\cB}{{\cal B}}
\newcommand{\cE}{{\cal E}}
\newcommand{\cG}{{\cal G}}
\newcommand{\cH}{{\cal H}}
\newcommand{\bcH}{\boldsymbol {\cal H}}
\newcommand{\cK}{{\cal K}}
\newcommand{\cM}{{\cal M}}
\newcommand{\cO}{{\cal O}}
\newcommand{\cR}{{\cal R}}
\newcommand{\cS}{{\cal S}}
\newcommand{\dcS}{\ddot{{\cal S}}}
\newcommand{\ds}{\ddot{{s}}}
\newcommand{\cT}{{\cal T}}
\newcommand{\cU}{{\cal U}}
\newcommand{\cV}{{\cal V}}
\newcommand{\wt}[1]{\widetilde{#1}}

\newcommand{\mA}{\mathbb{A}}
\newcommand{\mE}{\mathbb{E}}
\newcommand{\mG}{\mathbb{G}}
\newcommand{\mR}{\mathbb{R}}
\newcommand{\mS}{\mathbb{S}}
\newcommand{\mU}{\mathbb{U}}
\newcommand{\mV}{\mathbb{V}}
\newcommand{\mW}{\mathbb{W}}

\newcommand{\uq}{\underline{q}}
\newcommand{\ubq}{\underline{\boldsymbol q}}

\newcommand{\red}[1]{\textcolor[rgb]{0,0,0}{#1}}
\newcommand{\gre}[1]{\textcolor[rgb]{0,1,0}{#1}}
\newcommand{\blu}[1]{\textcolor[rgb]{0,0,0}{#1}}

\title{Imitation Learning-based Implicit Semantic-aware Communication Networks: Multi-layer Representation and Collaborative Reasoning}

\author{Yong~Xiao, \IEEEmembership{Senior~Member,~IEEE}, Zijian Sun, Guangming Shi, \IEEEmembership{Fellow, IEEE}, Dusit Niyato, \IEEEmembership{Fellow, IEEE}

\thanks{*This work has been accepted at IEEE Journal on Selected Areas in Communications. Copyright may be transferred without notice, after which this version may no longer be accessible.

(Corresponding author: Guangming Shi)

Yong Xiao is with the School of Electronic Information and Communications, Huazhong University of Science and Technology, Wuhan 430074, China, also with the Peng Cheng Laboratory, Shenzhen, Guangdong 518055, China, and also with the Pazhou Laboratory (Huangpu), Guangzhou, Guangdong 510555, China (e-mail: yongxiao@hust.edu.cn).

Zijian Sun is with the School of Electronic Information and Communications, Huazhong University of Science and Technology, Wuhan 430074, China (e-mail: jasons@hust.edu.cn).

G. Shi is with the Peng Cheng Laboratory, Shenzhen, Guangdong 518055, China, also with the School of Artificial Intelligence, the Xidian University, Xi’an, Shanxi, China, 710071, and also with Pazhou Lab (Huangpu), Guangdong 510555, China (e-mail: gmshi@xidian.edu.cn).

D. Niyato is with School of Computer Science and Engineering, Nanyang Technological University, 639798, Singapore (e-mail: dniyato@ntu.edu.sg).

Code is available at \texttt{https://github.com/zjs919/iRML}


}
}

\maketitle
\begin{abstract}
Semantic communication has recently attracted significant interest from both industry and academia due to its potential to transform the existing data-focused communication architecture towards a more generally intelligent and goal-oriented semantic-aware networking system. Despite its promising potential, semantic communications and semantic-aware networking are still in their infancy. Most existing works 
focus on transporting and delivering the explicit semantic information, e.g., labels or features of objects, that can be directly identified from the source signal. The original definition of semantics as well as recent results in cognitive neuroscience suggest that it is the implicit semantic information, in particular the hidden relations connecting different concepts and feature items that play the fundamental role in recognizing, communicating, and delivering  the real semantic meanings of messages.
Motivated by this observation, we propose a novel reasoning-based implicit semantic-aware communication network architecture that allows destination users to directly learn a reasoning mechanism that can automatically generate complex implicit semantic information based on a limited clue information sent by the source users. Our proposed architecture can be implemented in a multi-tier cloud/edge computing networks in which multiple tiers of cloud data center (CDC) and edge servers can collaborate 
and support efficient semantic encoding, decoding, and implicit semantic interpretation for multiple end-users. We introduce a new multi-layer representation of semantic information taking into consideration both the hierarchical structure of implicit semantics as well as the personalized inference preference of individual users. We model the semantic reasoning process as a reinforcement learning process and then propose an imitation-based semantic reasoning mechanism learning (iRML) solution to learning a reasoning policy that imitates the inference behavior of the source user.  A federated graph convolutional network (GCN)-based collaborative reasoning solution is proposed to allow multiple edge servers to jointly construct a shared semantic interpretation model based on decentralized semantic message samples. Extensive experiments have been conducted based on real-world datasets to evaluate the performance of our proposed architecture. Numerical results confirm that iRML offers up to 25.8 dB improvement on the semantic symbol error rate, compared to the semantic-irrelevant communication solutions. 
%
\end{abstract}

\begin{IEEEkeywords}
Semantic communication, implicit semantic-aware communication, multi-layer representation, multi-tier computing, collaborative reasoning, federated edge intelligence.
\end{IEEEkeywords}

\section{Introduction}
\label{Section_Intro}


Shannon’s information theory has been serving as the foundation for almost all existing communication systems.
In his seminal work published in 1948, Shannon observed that most communication messages involve semantic meaning. He, however, argued that ``these semantic aspects of communication are irrelevant to the engineering problem"\cite{Shannon1948}. 
Based on this argument, any given semantics-involving message must be first converted into a sequence of semantic-irrelevant binary symbols before being introduced into the physical layer communication process which is designed to focus only on transmitting and recovering messages with the bit-level accuracy. 

Recent development of communication and networking technology has witnessed a explosively growing demand on the smart services and applications, such as Augmented Reality/Virtual Reality/eXtended Reality (AR/ VR/XR), digital twins, and Tactile Internet, targeting at bringing human-like intelligence and immersive experiences into various aspects of human society\cite{XY2018TactileInternet, Yang2022NetMagazine, XY2021AdaptiveFog, XY2018EHFogComputing}. Most of these services are data-hungry, resource-consuming, and often require carefully orchestrated network resources and functionalities based on the users' background knowledge and experience as well as the semantics of the communication messages\cite{6Gpaper,XY2018UnlicencedNetSlice}. 

This motivates the semantic communication, a novel communication paradigm focusing on recognizing, delivering, and utilizing the key meaning of the messages during the communication and networking process\cite{XY2021SemComMagazine}. 
Recent development in neuroscience suggests that the semantic processing, including semantic recognition, communication, and inference, is a defining feature of human behavior, central not only to language, but also to human's capacity to exploit knowledge in learning, reasoning, and problem solving\cite{binder2009semantic}.
In other words, semantic communication 
has the potential to fundamentally transform the existing data-focused and semantic-irrelevant communication architecture towards a more generally intelligent and human need-driven semantic-aware networking system\cite{XY2021SemComMagazine,xiao2022ReasoningOnTheAir}.

The concept of semantic communication problem was first introduced by Weaver in 1949, right after Shannon introduced the classic information theory\cite{weaver1949recent}. In this work, the problems of communication  have been categorized into three levels and the Shannon theory has been coined as the solution of the level-one communication problem, referred to as the ``technical problem of communication". The semantic communication problem has been defined as the level-two problem which investigates ``how precisely do the transmitted symbols convey the desired meaning". Although Weaver also introduced the level-three problem, the ``effectiveness problem" investigating ``how effectively does the received meaning affect conduct in the desired way", he augured that the semantic and effectiveness problems are ``closely interrelated and overlapped in a rather vague way". In other words, the semantic communication has the potential to enable a higher-level meaning exchange with human-like semantic recognition, reasoning, and communication  that may revolutionize the way of human users' interactions with physical and virtual worlds\cite{xiao2022RateDistortion}.  

Most existing works consider the semantics of messages as the explicit semantics such as object labels and signal features that can be directly identified from the source signals\cite{Bourt2019SemanticImageTransmission,zhang2022UnawareTasks,Xie2020SemanticIoT}. It is known that the semantics involved in most communication messages can be much more than the explicit semantics. In fact, the original definition of semantics as well as recent results in cognitive neuroscience suggest that it is the implicit semantic information, in particular the hidden relations connecting different concepts, terms, features, and ideas, that plays a fundamental role in recognizing, communicating, and delivering the real semantic meanings between human users.
French philologist, Breal first defines the ``semantics" as the ``relationships between words and the knowledge they signify" in 1897\cite{breal1897essai}. Recent study in cognitive neuroscience suggests that the capability of human users to express rich semantic meanings based on very few words is achieved by combining 
words with different relations in different sequences. Furthermore, human users' high-level communication, learning, and inferring capability is also closely linked to their ability to 
establish relations between new unknown concepts and its known knowledge base. In other words, estimating and reasoning the possible relations to link the explicit semantics into hidden knowledge concepts  are of critical importance for enabling the high-level semantic communication and cognitive intelligence for the next generation human-oriented communication networking systems. 
%
%
%

Despite its importance, the development of the relation-based implicit semantic communication has been hindered by several challenges. First, communication messages may involve multiple types of complex relations connecting different subsets of knowledge concepts including both commonly-shared and fact-based concepts as well as private personal knowledge. It is generally impossible to establish and maintain a single database with all the global and private semantic knowledge. Second, the implicit semantics of messages may involve many hidden relations and knowledge concepts that 
cannot be directly observed from the source signal. This hidden information is often closely related to the message context,  users' personal preference, background, and private experience, many of which involve private information that is inaccessible for the destination user or any third-party service providers. Finally, relation-based implicit semantic information is difficult to represent, compress, recover, and evaluate. It may involve rich meaning 
information and attributes that are inefficient to transport and difficult to compress and compute. 
Currently, there is lacking a unified solution for representing, recovering and evaluating implicit semantics during the communication and networking process. 

In this paper, we investigate the implicit semantic communication from a novel perspective that is, instead of trying to maximize the detection and transportation of the explicit semantics, we propose an imitation learning-based solution to allow edge servers to learn from the past inference behavior of the user and establish a source user's background and preference-relevant reasoning mechanism to automatically infer the implicit semantics at the destination user. 
In this way, the destination user can imitate the past inference behavior of the source user and directly 
infer the implicit semantics from the observed clue information, e.g., explicit semantics, and also update and maintain the reasoning mechanism during the communication process. 
Motivated by recent observation in neuroscience suggesting that  the human brain utilizes a hierarchical structure of semantic knowledge for reasoning and processing disambiguity of semantic meaning\cite{Sarafyazd2019HierachcialReasoning}, we propose a new multi-layer representation of semantics, taking into consideration of both hierarchy of  semantics across different abstraction levels as well as personal preference of semantic reasoning for individual users.  
Inspired by the fact that human users tend to infer hidden information from the closely related concepts, we model the semantic reasoning process as a reinforcement learning process.
An imitation-based reasoning mechanism learning (iRML) solution is then introduced for the CDC and edge servers to learn a semantic reasoning mechanism, i.e., a reasoning policy, that can imitate the inference behavior of the source users. 
We also introduce a federated graph convolutional networks (GCN)-based  collaborative reasoning solution to allow multiple same-tier edge servers to collaborate in constructing semantic interpretation models without disclosing their local information.
Finally, extensive experiments have been conducted to evaluate the performance of our proposed solution based on real-world knowledge datasets. Numerical results suggest that iRML achieves up to 25.8 dB improvement on the semantic symbol error rate, compared to the existing semantic-irrelevant communication solutions.


We summarize the key contributions of this paper as follows:
\begin{itemize}
    \item[(1)] {\em Novel implicit semantic representation solution:} We propose a novel multi-layer representation of semantics including three key elements: explicit semantics, implicit semantics, and a user-relevant semantic reasoning mechanism. To improve the communication efficiency of the semantics, we 
    convert the rich semantics of messages into a sequence of low-dimensional semantic constellation representations that are efficient for physical channel transmission.  

    \item[(2)] {\em New imitation-based reasoning mechanism learning approach:} We propose a novel imitation-based reasoning mechanism learning solution, iRML, to allow edge servers and the CDC to learn from the user and train semantic reasoning mechanism models to imitate the inference behavior of the source users. To address the ill-posted problem of imitation learning, we adopt the principle of maximum causal entropy to convert the ill-posted problem into a strongly convex optimization problem. We then prove that in this case, each edge server can train a unique semantic reasoning mechanism with the minimized semantic distance between the generated interpretation and the expert reasoning paths observed by the source user.

    \item[(3)] {\em New multi-tier collaborative reasoning-based semantic interpretation solution:} We propose a federated GCN-based approach to allow the same-tier edge servers to collaborate in training a shared semantic interpretation model based on the decentralized knowledge datasets. Our approach does not require any edge server to expose its local knowledge information. We provide a theoretical bound on the convergence rate of the collaborative training process and also quantify the performance loss caused by the decentralized distribution of knowledge datasets.

    \item[(4)] {\em Extensive experimental results:} We evaluate the performance of our proposed architecture by simulating the implicit semantic-aware communication process based on real-world knowledge datasets. 
\end{itemize}

The remainder of this paper is organized as follows. 
Existing works that are relevant to semantic communication and knowledge reasoning are reviewed in Section \ref{Section_RelatedWork}. The multi-tier cloud/edge computing networks is presented in Section \ref{Section_Networkmodel}.
The multi-layer semantic representation and semantic distance are introduced in Section \ref{Section_SemRepresentationProblemForm}. The collaborative reasoning-based semantic communication solution is introduced in Section \ref{Section_MainAlgorithm}. We present numerical results in Section \ref{Section_NumericalResult} and conclude the paper in Section \ref{Section_Conclusion}.

\section{Related Work}
\label{Section_RelatedWork}

\noindent{\bf Multi-tier Computing:} Multi-tier computing is a promising architectural framework that seamlessly integrates CDC, fog, edge, and things for enabling diverse intelligent applications and services\cite{XY2012Stackelberg, XY2015MechanismDesign, chen2018fog,yang2019multi,Wang2022JointCachingMulti,Yang2021IoTBook}. Most of the existing works focus on optimizing the task scheduling and resource allocation to meet various application and service needs.  
For example, Chen et al. proposed a novel fog-as-a-service  architecture, called FA$^2$ST, that can offer end-to-end support for massive Internet-of-Things (IoT) systems\cite{chen2018fog}. Yang et al. introduced the multi-tier computing architecture and discussed its potential to deliver various IoT services in next generation mobile networks\cite{yang2019multi}. Wang et al. designed a hierarchical aggregation mechanism in a federated learning-based multi-tier network to further improve the utilization of resource efficiency\cite{wang2021resource}.  Wang et al. proposed a joint task offloading and caching optimization algorithm for massive MIMO-aided multi-tier computing networks\cite{Wang2022JointCachingMulti}.
In our paper, we apply multi-tier computing into collaborative reasoning-based semantic-aware communication networks. Our results suggest that multi-tier computing has the potential to serve as an important architectural framework for enabling semantic-aware network intelligence in the next generation communication networks. To the best of our knowledge, this is the first work that studies semantic-aware communication and its implementation under multi-tier computing networking systems.

\noindent{\bf Semantic Communication:} 
Most existing works in semantic communication can be categorized into two directions: information theory and machine learning.
Earlier works in semantic communication mainly focused on extending the Shannon's classic information theory to study semantic communication problem. For example, Carnap and Bar-Hillel introduced the semantic information theory in which the set of binary symbols in Shannon theory has been replaced with the set of possible models of worlds\cite{carnap1952outline}. Bao et al. derived the semantic entropy by simply replacing the Shannon's entropy of binary messages with the entropy of models\cite{Bao2011TowardsTheorySemanticComm}. Guler et al. investigated the coding error when the binary symbols have been replaced by words defined in a common dictionary-based dataset\cite{Guler2018SemanticCommGame}. It has been observed that most Shannon theory-based semantic communication work suffers from the so-called Bar-Hillel-Carnap (BHC) paradox which argues that any self-contradictory message has the maximum amount of information in Shannon theory due to its rarity, which contradicts with the fact that the semantic information must be true and contradictions are false with minimum meaningful information\cite{Stanford2022SemanticInfo}.   
Motivated by the observations that semantic information is learned and dynamically evolved through human interactions, in our recent study\cite{xiao2022RateDistortion}, we have developed the rate distortion theory for strategic semantic communication, a novel framework that combining game theoretic models with rate distortion theory to characterize the impact of interactions between semantic encoder and decoder on the distortion performance of the communication.

Motivated by the recent breakthrough in machine learning, especially the deep neural networks (DNNs)-based object/pattern recognition and classification algorithms, many recent works convert the semantic communication problem into the problems of recognizing and classifying the human-assigned labels and/or features of objects, called the explicit semantics, that can be identified from the various forms of source signals such as image\cite{LCChen2018ImageSemanticSegmentation,Noh2015DeconvolutionSemanticSegmentation,Ronneberger2015BiomedicalImageSegmentation}, voices\cite{Kiela2017LearningAudioSemantics,Santhanavijayan2021SpeechSemantic,Martin2021VoiceSemanticRecognition}, and text\cite{Tai2015ImprovedSemanticRepresentationLSTM,Dieng2016TextLongtermDependency,Kim2019SentenceSemanticMatching}, etc.  One of the key advantages of these solutions is that mature algorithms can be directly applied to identify semantics from the source signals. However, these solutions often require a large volume of manually labelled dataset for model training and updating. They also ignore the implicit impact such as those generated by interaction history, personal experience, and background on the semantics recognition and delivery process. 

Motivated by the recent study in cognitive neuroscience which observes that the human users are able to recover complex implicit semantics during the communication based on the background knowledge and/or a limited clue information, in this paper, we introduce the concept of implicit semantic-aware communication, in which implicit semantics including the hidden relations and inference mechanisms can be learned and automatically inferred during the communication and network interaction.
Different from the existing works, in the implicit semantic-aware communication,
instead of trying to maximize the total amount of information transmitted from one user to another, the source user will try to guide the destination user and/or the destination edge server to learn the implicit rules and reasoning mechanism that dominates the implicit semantic information generation process.
In this way, the BHC paradox can be naturally solved for rational users and also the personal preference and background knowledge of the source user is directly included into the learned mechanism. To the best of our knowledge, this is the first work that investigates the imitation learning-based implicit semantic-aware communication networks.   



\noindent{\bf Knowledge Reasoning:}
Knowledge reasoning has attracted significant interest recently due to its potential to infer hidden or even unknown information. Most of the existing works in knowledge reasoning are based on knowledge graph 
which can be roughly divided into three categories: graph-level, subgraph-level, and entity-level reasoning. Graph-level reasoning reveals the underlying properties of the entire input graphs taking into consideration their overall structural information. It has been successfully implemented in various real-world applications such as protein discovery\cite{yang2018learned} and molecule trials\cite{rong2020self}. Subgraph-level reasoning solutions  utilize relation features between knowledge entities exhibited in various subgraphs of knowledge\cite{yang2021consisrec}. Entity-level reasoning solutions utilize the entities' connection features and have shown promising results in entity classification and link prediction in various networking systems such as social networks\cite{zhou2008brief}. As mentioned earlier, knowledge graph can only model complete knowledge relations and therefore cannot be directly applied to represent the implicit knowledge components. In this paper, we proposed a novel multi-layer representation of semantics that can capture both explicit and implicit semantics. We also consider a collaborative reasoning-based solution that enables the collaborative learning among users decentralized knowledge bases involving both globally-shared and locally-owned knowledge information.  

\section{Multi-tier Cloud/Edge Computing Networks}
\label{Section_Networkmodel}
We consider a multi-tier cloud/edge network consisting of a number of edge servers deployed between the CDC and users as illustrated in Fig. \ref{Fig_NetworkModel}. The set of edge servers can be divided into different tiers according to their service types, coverage, as well as relative distances to the served users. To simplify our description, we focus on a three-tier network consisting of a CDC (high-tier) and a two-tier edge network including mid-tier and low-tier edge servers, to support the semantic communication and interpretation between users. 
Our considered network consists of the following key components: 

	\begin{figure}
		\centering
		\includegraphics[width=3.4 in]{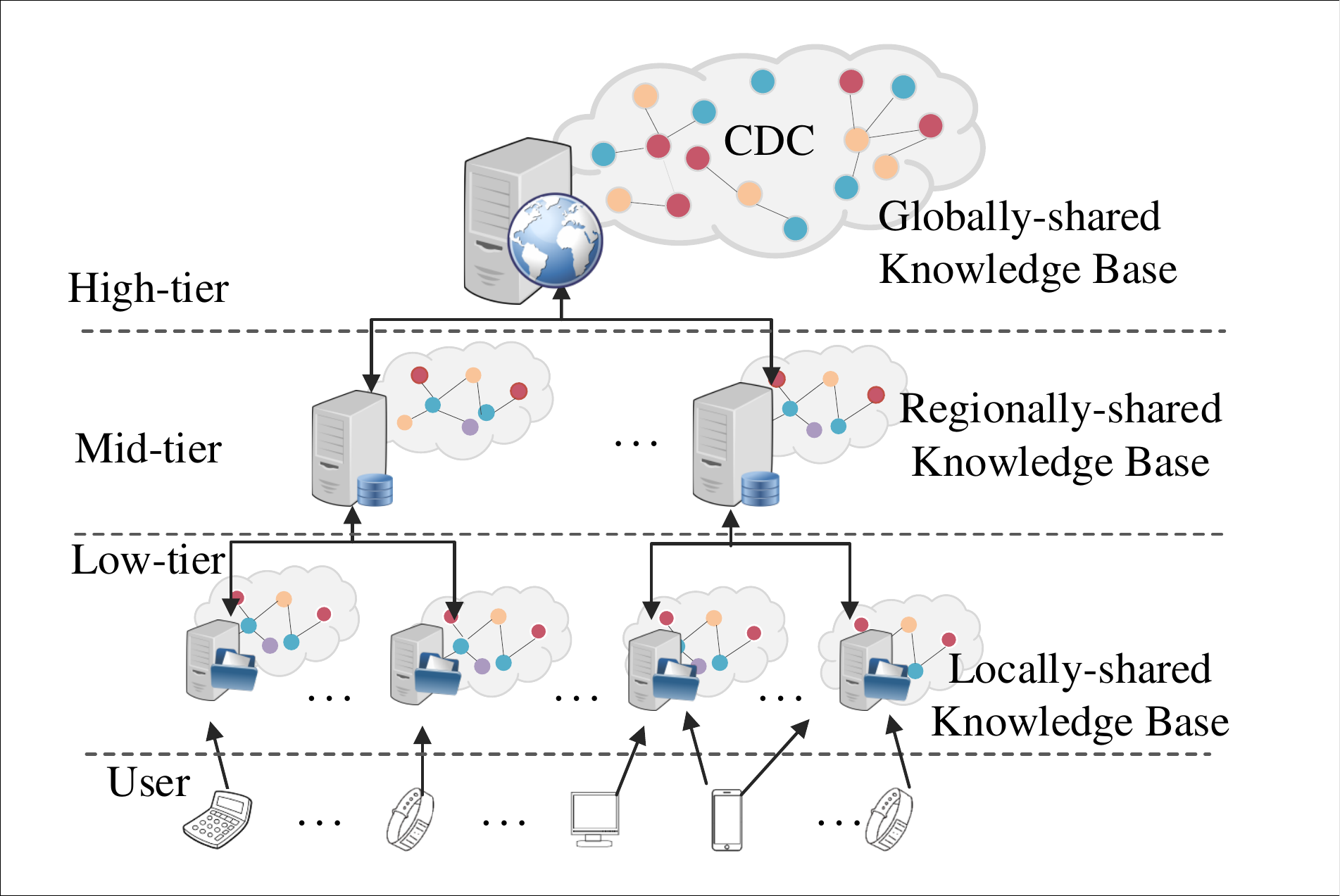}
		\caption{\footnotesize{A multi-tier cloud/edge computing network with globally/regionally/locally-shared knowledge bases.}}
		\label{Fig_NetworkModel}
	\end{figure}
	
\begin{itemize}

    \item[(1)] {\bf CDC:} corresponds to a centralized CDC that offers globally accessible computational and storage resources to all the users. The CDC can also maintain a globally shared semantic knowledge base consisting of accumulated knowledge entities, e.g., facts, terms, concepts, and objects, and possible relations, e.g., relationship between entities. In general, the entities and relations in the globally-shared knowledge base can correspond to the high-level commonly-shared facts that are irrelevant to the regional or local specific knowledge.     

    \item[(2)] {\bf Mid-tier Edge Servers:} correspond to the edge servers that provide regionally accessible computational and storage resources to the users in the coverage area. Similarly, each mid-tier edge server can also maintain a regionally shared knowledge base consisting of regionally-relevant and preferred knowledge facts, relations, customs, mechanisms, etc.  

    \item[(3)] {\bf Low-tier Edge Servers:} are local edge servers, each offers locally accessible computational and storage resources to the local users. Each local edge server also has a local knowledge base that can store some locally shared or even personalized knowledge including personalized and experience-based, e.g., biased understanding of knowledge concepts and relations associated with one or a limited number of individual users.

    \item[(4)] {\bf Users:} correspond to either information source and destination users that try to communicate their semantic meaning with each other. 
    We assume that each source user can observe the past communication history and extract a set of semantic reasoning trajectories consisting of globally, regionally, locally-shared, and/or even privately accessible semantic knowledge information. The source user will then use these observed semantic reasoning trajectories, called expert reasoning paths, to guide  
    CDC and edge servers across different tiers to train the semantic encoding, decoding, and interpreting models. 
\end{itemize}

Note that in our considered network architecture, the multi-tier CDC and edge servers are only required to be logically divided based on the privacy requirements and data and service coverage. For example, the function of a mid-tier edge server can be deployed within the CDC with an exclusive right to access the regionally shared  semantic knowledge, e.g., some users can purchase services offered by the CDC to store their private information data.  Similarly, mid-tier and low-tier edge servers can  also be co-located in the same device, e.g., edge server, with different rights to access the regional or local knowledge information. In some special cases, the mid-tier and low-tier edge servers can also be co-located with some high performance user devices with excessive computational and storage resources that can be utilized to store the regional or local knowledge. 

\section{Semantic Representation and Problem Formulation}
\label{Section_SemRepresentationProblemForm}

\subsection{Multi-layer Representation of Semantics}

As mentioned earlier, the semantic meaning of a message may involve both implicit and explicit semantic information that can be associated with different domains, privacy restrictions, and abstraction levels. Developing a simple and comprehensive way to represent the semantics of messages is therefore of critical importance for designing and implementing semantic communication systems.

Most existing works in semantic communications assume that all the possible semantic symbols are independent with each other and belong to a closed set available at every user. These representations of semantics contradict with the original definition of semantics introduced by Breal which defines the semantics as relationships between words and the associated knowledge components they represent. 

One potential solution to characterize the relations between concepts and knowledge components is to adopt the knowledge graph-like solutions to represent semantics of messages. 
Despite its high efficiency in representing complex relations between knowledge entities, applying graphical structures to represent semantic meaning of a message is also challenging due to the following reasons. 
First, knowledge graph can only represent a complete set of known entities and relations and therefore cannot represent the implicit meaning of messages. Second, most existing knowledge graphs are built based on facts and knowledge information recorded in globally-shared dictionaries carefully compiled by linguists. It cannot reflect the local preference of inference or some personalized understanding, e.g., biased personal experience-based  knowledge, of  users. In other words, it is challenging to apply a single knowledge graph to include both the globally-shared knowledge as well as some regionally-shared or privately-owned information of individual users. Finally, some  messages may consist of  knowledge concepts and relations associated with multiple decentralized knowledge bases deployed across multiple CDC and/or edge servers. This makes it difficult for performing joint semantic inference and interpretation across multiple knowledge bases without causing the leakage of private information. 


\begin{figure}
\centering
    \includegraphics[width=3.4 in]{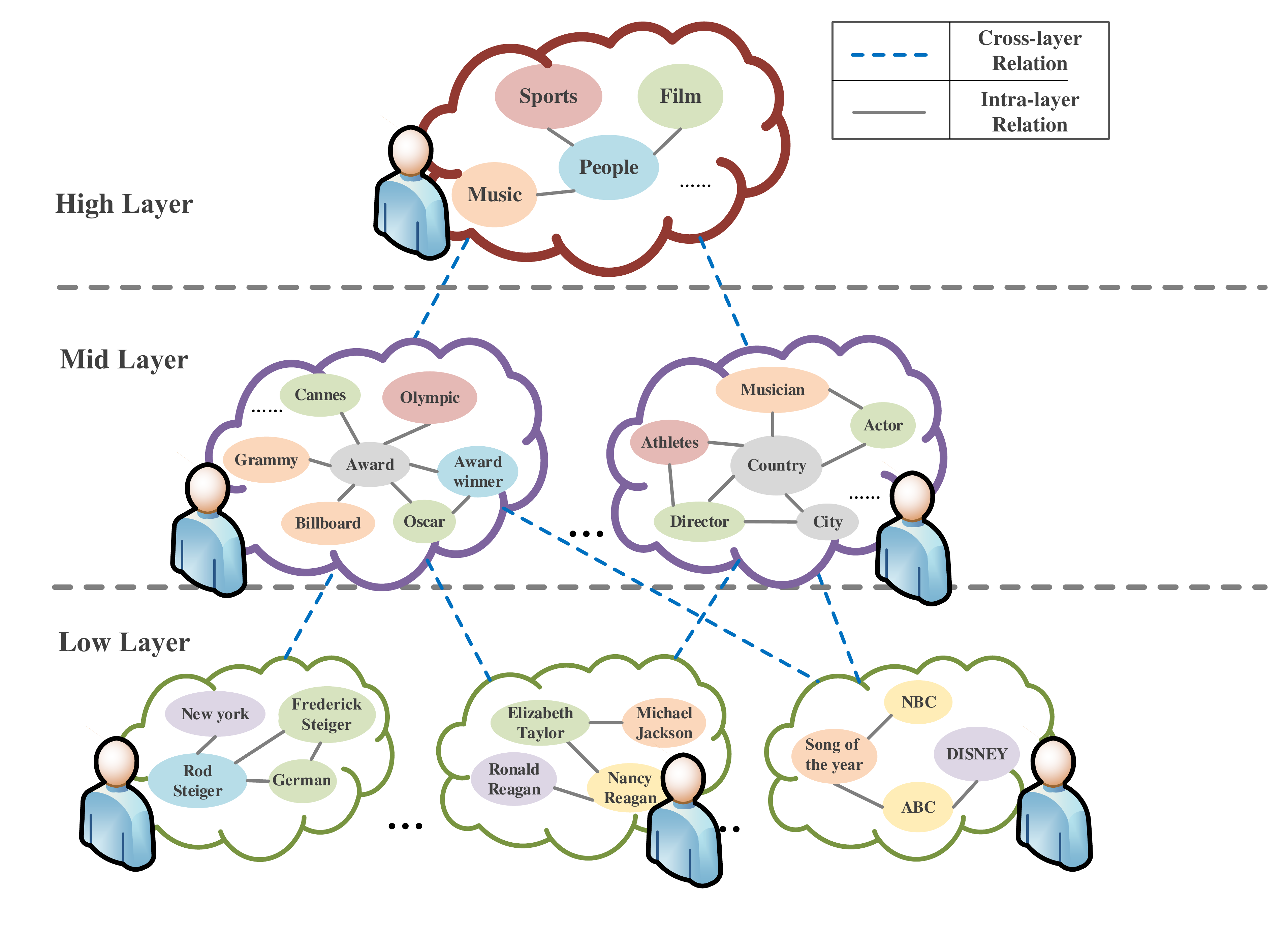}
    \label{Fig_MultiConcepts}
\caption{\footnotesize{Knowledge entities of a dictionary-based dataset (FB15K-237) divided into three layers of abstraction.}} 
\label{Fig_MultiLayerKnowledgeRepresentation}
\end{figure}

Recent study in cognitive neuroscience reports that the semantic cognitive process of  human users can be considered as a multi-layer semantic reasoning and causal inference process, called hierarchical reasoning\cite{Sarafyazd2019HierachcialReasoning}, consisting of multiple abstraction layers of the cognitive process from the highest layer, involving the high level abstraction of concepts and the cognitive process, extending to the lowest layer, often associated with very concrete and simple information and cognitive actions.
%

Motivated by the above study, in this paper, we propose a novel multi-layer representation of semantics of messages taking into consideration 
both cross-layer hierarchical structure of knowledge reasoning process across different abstraction layers of knowledge components and intra-layer network connections between the closely related entities with similar abstraction layer. In particular, we seek interpretations of semantics of a given message in $L$ different abstraction layers where the root layer (highest) consists of high-level conceptual level entities and relations such as domain names and classes and/or types of knowledge entities 
and the other layers are composed of hyponymy concepts and relations extended from one or multiple parent layer entities with more concrete knowledge information. 
The lowest layer consists of highly detailed entities and relations such as location points and specific names of people. 
%
In fact, we can observe the  hierarchical structure of knowledge entities and relations in many real-world human knowledge datasets. 
In the dictionary-based knowledge dataset, e.g., FB15K-237\cite{toutanova2015FB15k237}, for example, 
the knowledge entities can be categorized into different abstraction layers based on the rank of their degrees (numbers of directly connected relations), as illustrated in Fig. \ref{Fig_MultiLayerKnowledgeRepresentation}. 
In particular, if we divide all the entities in the FB15K-237 into three different layers based on the rank of their degree: high layer (entities with degrees above 50), middle layer (entities with degrees between 6 and 50), and low layer (entities with degrees less than 6). We can observe that the knowledge entities in the highest layer correspond to concepts with the highest level of abstractions such as ``sports", ``film", 
``people", 
and ``music". Although the number of high layer entities only accounts for a small portion, less than 0.0814\%, of the entire knowledge base, they are often closely related to each other, for example, people and sports are often the main themes of films. 
The middle layer entities are knowledge concepts with relatively more detailed knowledge terms such as ``film awards", ``music awards", 
and ``country". 
Each middle layer entity can be associated with multiple concepts in the high layer, for example, ``Oscar award" contains both best ``musical award" and best ``film award". Generally speaking, the entities in the  middle layer are less connected to each other compared to  those in the high layer. 
The low layer entities offer the most concrete information compared to the middle layer entities, e.g., ``Ronald Reagan" and ``Elizabeth Taylor", etc. Similarly, each entity in the low layer can be correlated with multiple entities in the middle layer, e.g., ``Ronald Reagan" is an actor and also a president of a country.

More formally, we define a {\em multi-layer  representation of semantics of a message} as 
a triple $\bomega = \langle \bv^E$, $\bp^{(L)}\left( \bv^E \right), \bpi \rangle$ 
consisting of the following elements:
\begin{itemize}
\item[(1)] {\bf Explicit Semantics:} include the visible entities (concepts, objects, terms, etc.) and relations (relationship between entities) that can be directly identified from the source signal. Let $\bv^E = \langle \be^E, \br^E \rangle$ be the explicit semantics of a given message where $\be^E$ and $\br^E$ correspond to the identifiable entities and relations, respectively.

\item[(2)] {\bf Inferred (Implicit) Semantics:} correspond to the implicit knowledge components including the hidden entities and relations that are closely related to the semantic meaning of the message. 
Motivated by the fact that human users tend to infer implicit knowledge components from their closely related concepts and relations, we model the inference process of semantics as a sequential decision making process which extends one semantic relation or entity at a time from the set of explicit semantics $\bv^E$. For example, an inferred semantic path extended from a visible entity $e_0$ after $t$ sequential inference processes is given by $p_t=\langle e_0, r_1, e_1, r_2, e_2, \ldots, r_t, e_t \rangle$ where $e_0 \in \be^E$ is a visible entity identified in the source signal and $r_1, e_1, r_2, \text{ and } e_2$ are implicit semantic entities and relations extended from $e_0$. 
Since the semantics of messages can be inferred based on entities and relations across various abstraction levels, we use $\bp^l \left( \bv^E \right)$ to denote the set of paths extended from the explicit semantics to the $l$th abstraction level of the knowledge entities and relations. 
Since in this paper, we focus mainly on minimizing the semantic disambiguity (relatively detailed meaning) of messages, we only consider the semantic reasoning of any given entity in its downward layers, i.e., for a layer $l$ entity $e_0$, it will only infer implicit semantics in the layers that are equal to or lower than layer $l$. Suppose $L$ abstraction layers are ranked from the highest to the lowest from 1 to $L$. Each entity $e_0$ will generate $(L-l+1)$ paths in abstraction layers $l, l+1, \ldots, L$, e.g., we can write the set of reasoning paths generated from $e_0$ as $\bp^{(L)}\left( e_0 \right) = \{\bp^{i}\left( e_0 \right)\}_{i\in {\{l, \ldots, L\}}}$.
We also write the combination of semantic inference paths for all the explicit semantics $\bv^E$ as $\bp^{(L)}\left( \bv^E \right) = \{\bp^{L}\left( e_0 \right)\}_{e_0\in {\bv^E}}$.

\item[(3)] {\bf Inference Mechanism:} corresponds to the inference rules that decide the potential connections between the explicit and implicit semantics. As mentioned earlier, the implicit semantics cannot be directly obtained from the explicit semantics, but will have to be inferred based on the background and/or personally related information such as personal preference and previous experience. 
For example, a message ``Micheal is reading a book about Tesla" consists of key entity ``Tesla" which can be closely linked to either the electric vehicle manufacturer or the inventor Nikola Tesla, none of which can be directly interpreted from the received message itself. If recent conversations between Micheal and his friends mentioned some inventions during 1890s, it is most likely that ``Tesla" in the example message is referred to as the ``inventor Nikola Tesla". In other words, an inference mechanism can be considered as a user-related mapping function that maps the observed explicit semantics into a set of possible reasoning trajectories extended across different abstraction layers, i.e., we can write the inference mechanism as $\bpi: \bv^E \rightarrow \bp^{(L)}$. In this paper, we assume that all the communication messages arrived at the users are generated by an unknown and unobserved inference mechanism, referred to as the expert inference mechanism. We then focus on developing solutions to 
learn an estimated inference mechanism to imitate the true expert inference mechanism from a set of reasoning trajectories observed during the past communication history.



\end{itemize}

Note that in the previous section, we divide the knowledge entities and relations into different tiers according to the privacy requirements and service coverage of edge servers. In this section, however, we consider categorization of knowledge entities as well as their associated relations based on the abstraction levels. These two ways of division of knowledge entities do not have to be correlated with each other. In other words, different mid or low-tier edge server can have knowledge entities associated with different combinations of abstraction levels.


\subsection{Semantic Distance}
\label{Subsection_SemDistance}
One of the key objectives of semantic commutations is to minimize the semantic distance, a metric for measuring the meaning dissimilarity between the true meaning of the source user and the recovered meaning interpreted by the destination user. 
%
Unfortunately, designing a simple and unified metric to measure semantic distance between two semantic representations is known to be a notoriously difficult task due to the following reasons. First, the representations of semantics involve complex relations and entities at different abstraction levels, there is still lacking a unified  metric that can characterize the difference between two hierarchical structures of semantic meanings. Second, since the real intended meaning, especially the implicit semantic meaning, of the source users is generally unknown or difficult to recognize, evaluating the dissimilarity between an interpreted meaning and the true meaning of the user is generally impossible. 
Finally, as mentioned earlier, the implicit semantics need to be inferred based on some background and personality-related information that cannot be directly observed from the source signal. In other words, different users may have different and even biased understanding of meanings when observing the same explicit semantics. How to capture the impact of the personal-related information on the semantic distance between different interpretations is still an open problem.   

In this paper, we propose a unified solution for measuring the distance of any given pair of semantic representations. As mentioned previously, \blu{the semantics of a message can be represented by a set of implicit semantic reasoning paths generated from the observed explicit semantics. Suppose the source user can observe a history of communication messages consisting of many previously observed semantic reasoning trajectories of knowledge entities and relations originated from some commonly used entities. We refer to these semantic reasoning trajectories as the {\em expert reasoning paths}. Generally speaking, the communicated messages of a specific user tend to follow the similar syntax and reasoning preference. The expert reasoning paths can therefore be considered as the random samples of an unknown semantic inference mechanism of the source user.} 
Let $\bq \left( \bv^E \right)$ be the set of expert reasoning paths originated from the visible entities and relations in $\bv^E$, which follows a stationary distribution under the given $\bv^E$. We also use $\bp \left( \bv^E \right)$ to denote the interpreted semantic paths made by the destination user or its associated edge server when observes $\bv^E$. Since the semantic reasoning mechanism is closely related to user's background and personal preference and is generally assumed to be stationary for the same user, we can then define the semantic distance between the semantic meaning of the source user and that interpreted by the destination user as the difference between the distributions of two semantic reasoning paths generated by their corresponding reasoning mechanisms based on the same explicit semantics $\bv^E$, denoted as $\Gamma \left( \bq\left( \bv^E \right), \bp\left( \bv^E \right) \right)$.   

As mentioned earlier, the reasoning process of human users tend to follow a sequential inference process dominated by a reasoning mechanism, a policy that infers the most relevant hidden relations and semantic entities from the observed or previous inferred knowledge components. 
In particular, we can write  
the reasoning policy $\bpi_{E}$ that generates the expert paths as a mapping function mapping the last entity of a reasoning path into the next possible hidden relations, i.e., $\bpi_{E}: \bp_t \rightarrow \br_{t+1}$ for $t=0, 1, \ldots$.
We define the {\it occupancy measure} of a reasoning policy $\bpi_E$ as the probability of observing a set of relations $\br$ being added to a set of paths $\bp$ when the user infers implicit semantics based on the policy $\bpi_E$, i.e., we can write the occupancy measure of policy $\bpi_E$ as $\bc_{\bpi_E} \left( \br, \bp \right) = \bpi_E\left( \br, \bp \right) \Pr\left( \bp|\bpi_E \right)$ where $\bpi_E\left( \br, \bp \right)$ is the probability of selecting relation $\br$ under path $\bp$ decided based on policy $\bpi_E$. If we assume that the reasoning process follows the Markov property, that is the current choice of relations $\br$ only depends on the previously observed or inferred entities $\be$, we can rewrite the occupancy measure of policy $\bpi_E$ as $\bc_{\bpi_E} \left( \br, \be \right) = \bpi_E\left( \br, \be \right) \Pr\left( \be|\bpi_E \right)$. We can observe that for any given explicit semantics, the resulting semantic reasoning paths can be fully determined by the reasoning policy with the maximum length constraint $J$ of a reasoning path. In other words, the semantic distance between the original meaning and the interpreted meaning is, in fact, the difference between the reasoning mechanisms of the source and destination users, characterized by the occupancy measures of their generated paths. Let $\bpi_D$ be the reasoning mechanism learned by the destination users. Let $\bq_{\bpi_E}$ and $\bp_{\bpi_D}$ be the paths generated by reasoning mechanisms of the expert reasoning mechanism and that generated by the reasoning mechanism learned by the destination user based on our proposed solution, respectively.
%
In the rest of this paper, we mainly focus on two types of distance metrics for measuring semantic distances of reasoning mechanisms defined as follows. Our proposed solutions, however, can be extended into more general scenarios involving other forms of semantic distances. 
\begin{itemize}
    \item[(1)] {\bf Statistic-based Semantic Distance (Distance-I)}: The semantic distance between the expert paths and inferred paths can be measured by their statistic difference. For example, suppose that the set of all the valid occupancy measures 
    of the expert paths and that of the interpreted paths are given by $\Delta^E$ and $\Delta^D$, respectively, and if we adopt the cross-entropy, one of the most commonly used metrics for measuring the statistic difference, the semantic distance can then be written as
\begin{eqnarray}
\Gamma \left( \bq_{\bpi_E}, \bp_{\bpi_D} \right) = \mE_{\bc_{\bpi_E} \sim \Delta^E} [{-\log \left(\bc_{\bpi_D}\right)}]. 
\label{eq_comparator}
\end{eqnarray}

    \item[(2)] {\bf Energy-based Semantic Distance (Distance-II)}: We also consider an energy-based solution to first project the 
    high-dimensional graphical representation of reasoning trajectories 
    into a low-dimensional space, called semantic space, in which the  dissimilarity between two semantic meanings is proportional to their Euclidean distance. In particular, let $\tilde{\be}$ and $\tilde{\br}$ be the representations of $\be$ and $\br$ in semantic space, respectively. The main objective is to design a projection function such that the Euclidean distance between $\tbe_t+\tbr_{t+1}$ and $\tbe_{t+1}$ is minimized if entities $\be_t$ and $\be_{t+1}$ are connected via relation $\br_t$. Suppose $J$ is the maximum length of each semantic reasoning path. In this way, we can write the representation of a semantic reasoning path in the semantic space as $\tbp=\sum\limits^J_{j=1} \tbr_j + \tbe_0$.  The semantic distance between reasoning policies $\bpi_E$ and $\bpi_D$ will then be defined as:
    \begin{eqnarray}
    \Gamma\left(\bq_{\bpi_E}, \bp_{\bpi_D} \right) &=& {\mE_{\bc_{\bpi_E}}} {\|\tbe_t+\tbr_{t+1}-\tbe_{t+1}\|^2} \nonumber\\
    &&- \mE_{\bc_{\bpi_D}} \|\tbe_t+\tbr'_{t+1}-\tbe'_{t+1}\|^2,
    \end{eqnarray}
    where $\tbr'_{t+1}$ and $\tbe'_{t+1}$ are the hidden relations and entities inferred by $\bpi_D$. 

\end{itemize}

\blu{The above two distance metrics have different features and can be applied into different scenarios. In particular, the energy-based semantic distance is simple to calculate and easy to scale into large knowledge base. Also, since it measures the semantic difference (e.g., plausibility of true facts compared to the false ones) based on the Euclidean distance in the projected semantic (encoding) space, it is also easier to evaluate the noise combating performance as will be discussed later in this paper. The statistic-based semantic distance needs to be calculated based on the probability distributions of possible semantic meanings and therefore is more suitable to measure the semantic difference between semantic generation rules or models, i.e., the semantic reasoning paths generated by the neural networks, especially the deep neural networks-generated paths.} 

\subsection{Problem Formulation}

The main objective of this paper is to design a semantic-aware communication solution to minimize the semantic distance between the true meaning of the source user and the interpreted meaning of the destination under any given explicit semantics.
We can write the optimization problem as follows:
\begin{eqnarray}
\min_{\bpi_D} \mathbb{E}_{\bpi_D} \left( \Gamma\left(\bq_{\bpi_E},\bp_{\bpi_D} \right) \right) 
\label{eq_MainProblem1},
\end{eqnarray}
where $\bpi_E$ is the unobserved true semantic reasoning mechanism of the source user.


It can be observed that problem (\ref{eq_MainProblem1}) is an ill-posted problem which is generally difficult to solve due to the following reasons. 
First, the users cannot directly access the reasoning mechanism that generates the expert paths but can only observe a limited set of trajectories of the knowledge components observed during the past communication. Considering that there exist multiple different ways/paths to express the same/similar semantic meaning, how to estimate the reasoning mechanism that imitates the true reasoning process of the source user is a challenging task. Second, different edge servers may access different sets of decentralized knowledge bases and therefore how to develop a resource efficient knowledge sharing framework without compromising the privacy of the knowledge information is also an open problem.

\section{Collaborative Reasoning-based Semantic-aware Communication Network Architecture}
\label{Section_MainAlgorithm}
\begin{figure}[!ht]
\centerline{\includegraphics[width=3.4 in]{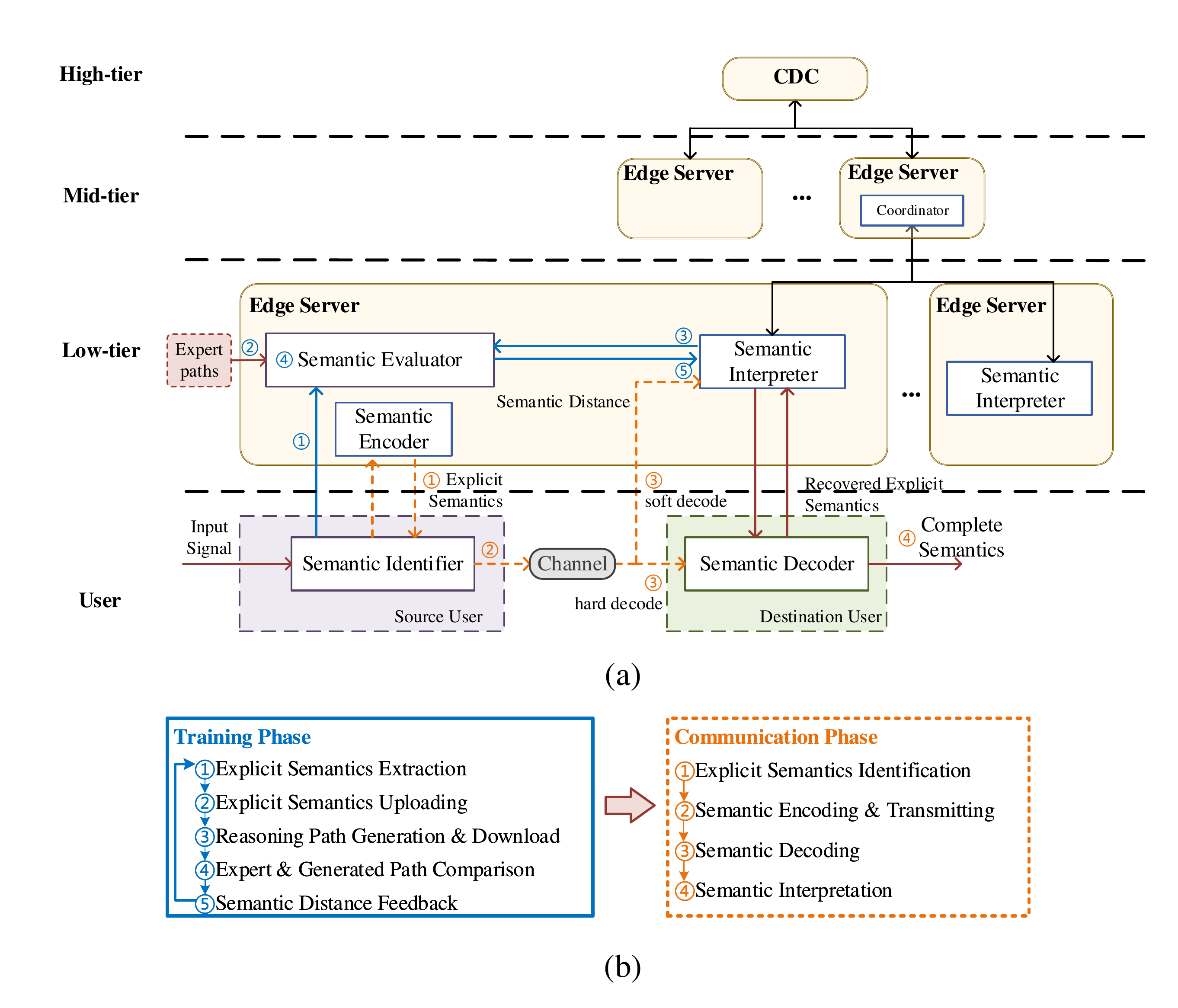}}
\caption{\footnotesize{(a) Collaborative reasoning-based implicit semantic-aware communication network architecture, and (b) procedures in the training and communication phrases.}}
\label{Fig_CRSAC}
\end{figure}

\subsection{Architecture Overview}
We propose a collaborative reasoning-based implicit semantic-aware communication network architecture as illustrated in Fig. \ref{Fig_CRSAC}. There are two phases to implement the proposed architecture: {\em (Model) training phase} and {\em (semantic) communication phase}. \blu{In the training phase, each source to destination user pair assists the CDC and edge servers to train semantic encoding and interpreting models. In particular, the source user will first identify some initial entities and/or relations from the expert reasoning paths as the explicit semantics to be sent to the semantic interpreter at the edge server that is close to the destination user. The semantic interpreter at the destination user will then generate a set of possible semantic reasoning paths to be sent back to the semantic evaluator at the edge server of the source user. The source user will then compare the  paths generated by the semantic interpreter with the expert reasoning paths and feedback the value of semantic distance to the semantic interpreter. The above process will be repeated until the semantic interpreter converges to a stationary policy and semantic evaluator at the source user cannot differentiate the paths generated by the semantic interpreter from expert reasoning paths. The edge server will also train a semantic encoder to convert the high dimensional representation of explicit semantic entities and relations into a set of low-dimensional semantic representation that is efficient for physical channel transmission. The trained semantic encoder will be loaded to the source user for message encoding during the semantic communication phase.}
 Similarly, to train the semantic decoder, the destination user will upload the  noisy version of the low-dimensional semantic signal received from the channel to the edge server. The edge server can then calculate a decoding function that can recover the semantics of the source user. The semantic decoder will also be loaded to the destination user at the end of the training phrase. \blu{Note that, during the training process of the semantic-aware communication system, the semantic interpreter at the destination user uses the noisy version of the explicit semantics as input and can then output the inferred implicit semantics based on the noisy semantics. In other words, if the destination user communicates with the semantic interpreter at the edge server via a high SNR channel, e.g., wired connection, we can ignore the noise of channel connecting edge servers and users. However, if the destination user connects the edge server through wireless channel, the signal arrived at the edge servers will include both noises in the channel between the source and destination users as well as that connecting the destination user and the edge server. In this case, the semantic interpreter will use different input to train the semantic interpretation model and the rest of the training process remains the same.}
We will discuss the semantic encoding and decoding process for a single edge server case in Section \ref{Subsection_iRMLSingleEdgeServer}. Multiple edge servers can collaborate in training their semantic interpreters using a federated GCN-based collaborative reasoning solution which will be discussed later in Section \ref{Subsection_iRMLCollaberativeReasoning}.

During the communication phrase, the source user will first identify the explicit semantics and then apply the pre-loaded semantic encoder to compress the identified semantics for physical channel transmission. The destination user once recovers the explicit semantics from the received signal will then send the recovered semantics to the semantic interpreter at the edge server. The semantic interpreter will generate the implicit semantics to be sent to the destination user.

\subsection{Semantic Reasoning Mechanism Modeling}

Motivated by the fact  that, in most human-related communication scenarios, the current communication context is closely related to the communication history, in this paper, we model the sequential semantic reasoning process as a reinforcement learning problem in which the implicit semantic entities and relations are sequentially inferred from the explicit semantics. In this way, the reasoning mechanism can then be considered as an inference policy that decides the possible (hidden) relations that link the currently interpreted entities to the next relevant hidden entities.  

We focus on a slotted sequential decision making process in which, in each time slot $t$, the user decides the possible relations to extend the current reasoning paths. We use subscript $t$ to denote the parameters decided in the $t$th reasoning iteration for $0\le t\le J$. We define the semantic reasoning process of a user as a four-tuple $\langle {\cal S}, {\cal A}, \bpi, {\mu} \rangle$, where the state space ${\cal S}$ corresponds to the currently extended reasoning paths from the explicit semantics, e.g., we can write $s_t = \langle \bp^{(L)}_t \left( \bv^E \right), t \rangle$ where $\bp^{(L)}_t = \langle e^l_0, \ldots, e^l_t \rangle_{l\in \{1, \ldots, L\}}$. We have $s_0=\bv^E$. The action space ${\cal A}$ is the possible implicit relations selected to extend the current reasoning paths, i.e., action $a_t$ decided in the $t$th time slot is given by $a_t = \br_t$. The policy $\bpi$ corresponds to a set of stationary stochastic policies that decide the actions under each given state, i.e., $\bpi: \cS\times \cA \rightarrow \cA$. It can be observed that for any given explicit semantics (initial state $s_0$) and a policy $\bpi$, the set of possible reasoning paths $\bq$ (a sequence of actions) can be determined. 
We can therefore write the set of expert paths generated by policy $\bpi_E$ with explicit semantics $\bv^E$ as $\bq_{\bpi_E} \left( \bv^E \right)$. 

\subsection{Imitation-based Reasoning Mechanism Learning}
\label{Subsection_Interpreter}
Since the expert policy $\bpi_E$ is unknown and cannot be observed by the user, how to estimate the policy function to imitate the true reasoning process of the source user is a very challenging task. One commonly used approach is to adopt the inverse reinforcement learning-based solutions which  
need to first estimate a specific cost function and then applies the traditional reinforcement learning-based solutions to calculate the policy. One of the key assumptions of these solutions is that every expert path is assumed to be a unique and optimal choice of the user to minimize its cost. In many practical systems, however, it is impossible to estimate any specific cost function. Also, due to the randomness of users' choices, it is unrealistic to assume that every reasoning decision made by the users is optimal.

In fact, each of the  expert paths observed by the source user can be considered as a sample resulted from many different possible probability distribution functions, due to the lack of information, none of which can be assumed to be more likely than others. This motivates a novel theoretical framework, called the principle of maximum entropy, which resolves the uncertainty about some constrained probability distribution functions for ``modeling the observed behavior of the users by choosing the distribution functions that has the least commitment for any particular outcome"\cite{Ziebart2013MaxCausalEntropy}. In other words, a probability distribution function that follows the  principle of maximum entropy is the distribution that maximizes the entropy, a standard measure of uncertainty in information theory, subject to the matching of observed behavior trajectories.

In this paper, our main objective is to learn a reasoning policy that captures the sequential decision making rules dominated by the causality between the explicit and implicit semantics, e.g., any hidden entities or relations should only be inferred during the interpretation of semantic meaning when they are closely related to the other existing semantic components of the arrived message. Therefore, we adopt the principle of maximum causal entropy, an extension of principle of maximum entropy that involves the conditional probability distributions for capturing the sequential decision making dominated by the causality between sequence of actions, to learn the reasoning mechanism that explains the observed reasoning paths of the source user. More formally, the casual entropy of a state to action transition is defined as $H\left( a_t \| s_t \right) = \mathbb{E}_{a|s} \left[ -\log \left( \Pr\left( a_t | s_{1:t}, a_{1:t} \right) \right)\right]= \mathbb{E}_{\bpi} \left[ - \log \bpi \left( a,s \right) \right]=H(\bpi)$ where $H\left( \cdot \right)$ is the entropy function, $s_{1:t} = \langle s_l \rangle_{l\in \{1, \ldots, t\}}$ and $a_{1:t-1} = \langle a_l \rangle_{l\in \{1, \ldots, t-1\}}$. The main idea of the principle of maximum causal entropy is to find a policy that maximizes $H(\bpi)$.

We can now propose an imitation learning solution based on the principle of maximum entropy to train a reasoning policy to match the user observed semantic reasoning process.
%
Let us first consider an imitation learning-based solution for a single edge server to learn the semantic reasoning policy of the user. We extend our proposed imitation learning solution into a large number of multi-tier collaborative edge severs in the next subsection.
In the rest of this subsection, we first introduce our proposed solution and then prove that the proposed solution can, in fact, address all the issues of the ill-posted problem of (\ref{eq_MainProblem1}).

Since the user cannot know the reasoning policy of the source user but can only observe expert reasoning trajectories, we try to derive a reasoning policy than can match the state to action mapping observed in the expert paths. Let us define the occupancy measure of a given policy $\bpi$ as the probability of observing an action being decided under a given state when the user reasons based on policy $\bpi$, i.e., the occupancy measure of policy $\bpi$ is defined as $\bc_{\bpi} \left( a, s \right) = \bpi\left( a, s \right) \Pr\left( s|\bpi \right)$ where $\bpi\left( a, s \right)$ is the probability of selecting action $a$ under state $s$ under policy $\bpi$. Based on this definition, for a given occupancy measure $\bc$, we can rewrite policy $\bpi$ as $\bpi_c\left(a, s \right)={\bc(a, s)/\sum_{a'} \bc(a', s)}$. In this way, we can convert the matching between semantic reasoning policies $\bpi$ and $\bpi_{E}$ as the matching between occupancy measures $\bc_{\bpi}$ and $\bc_{\bpi_E}$.  
We can finally write the maximum causal entropy problem with the constraints on the matching between the semantic reasoning policy $\bpi_D$ learned by the edge server and the true reasoning mechanism $\bpi_E$ that generates the expert path as follows:
\begin{eqnarray}
&& \max\limits_{\bpi_D} \;\;\; H \left( \bpi_D \right)
\label{eq_MaxCauEntropyProblem}\\
&&\;\; \mbox{s.t.}\;\;\;\; \bc_{\bpi_D} \left( s, a \right) = \bc_{\bpi_E} \left( s, a \right), \forall s \in {\cal S}, a \in {\cal A} \nonumber 
\end{eqnarray}
where $H\left( \bpi_D \right) = \mathbb{E}_{\bpi_D} \left[ - \log \bpi \left( a, s \right) \right]$ is the causal-entropy of the semantic reasoning policy $\bpi_D$.

By substituting the definition of semantic distance defined in Section \ref{Subsection_SemDistance} and convert the above problem into its Lagrangian form, we can have the following optimization problem,  
\begin{eqnarray}
\min\limits_{\bpi_D} \;\;\; F \left(\bpi_D\right) 
\label{eq_MaxLagCauEntropyProblem}
\end{eqnarray}
where $F$ is the loss function given by
\begin{eqnarray}
&& F \left(\bpi_D\right) \nonumber \\
&& =\; - H \left( \bpi_D \right)  + \lambda \left( \mathbb{E}_{s, a}  \bc_{\bpi_D} \left( s, a \right) - \mathbb{E}_{s, a} \bc_{\bpi_E} \left( s, a \right) \right) \nonumber \\
&& =\; - H \left( \bpi_D \right)  + \lambda \Gamma\left( \bpi_E, \bpi_D \right),
\end{eqnarray}
\blu{where $\lambda$ is the Lagrange multiplier. Since $F \left(\bpi_D\right)$ is convex, we can apply the standard convex optimization solution to solve problem (\ref{eq_MaxLagCauEntropyProblem}).\cite{li2009lagrange} }

We can now prove that by converting problem (\ref{eq_MainProblem1}) into problem (\ref{eq_MaxCauEntropyProblem}), the problem is no longer ill-posted and we can find a unique optimal solution by applying the standard convex optimization solution. 

In particular, we have the following results:

\begin{theorem}\label{Theorem_Convexity}
Suppose occupancy measure $\bc$ is non-negative and its value is bound by a positive constant $M$, i.e., $0 \leq \bc \leq M$. There exists a constant $\xi \in (0,1)$, such that the corresponding policy $\bpi_c\left(a, s \right)={\bc(a, s)/\sum_{a'} \bc(a', s)} \leq 1-\xi$. Then, problem (\ref{eq_MaxLagCauEntropyProblem}) is strongly convex and there exists a unique policy under the observed occupancy measure.
\end{theorem}
\begin{IEEEproof}
See Appendix \ref{Proof_Convexity}.
\end{IEEEproof}

From the above theorem, we can observe that by regularizing the loss function based on the causal entropy, we can always find a unique semantic reasoning policy that imitates the reasoning behavior of the source user.



\subsection{iRML with a Single Edge Server}
\label{Subsection_iRMLSingleEdgeServer}
In our considered implicit semantic-aware communication networks, each user may not have the sufficient resource to perform semantic coding but can guide the resource-abundant edge servers to learn their personalized semantic reasoning preference. We propose a novel solution, iRML, for training both semantic encoder and decoder at CDC and edge servers to support semantic communication between a source-to-destination user pair. Let us first introduce the training process of iRML for a single edge server scenario consisting of the following key procedures. 

\subsubsection{Training Semantic Encoder at the Edge Server}
The source user will try to train a semantic encoder for converting the high-dimensional semantic representation of messages into a low-dimensional representation that is efficient for physical channel transmission. Different from the traditional encoding solution which converts messages into a sequence of binary symbols regardless of their represented meaning, 
in our proposed semantic-aware communication, different entities are encoded into a sequence of semantic constellation representations in which the distance between entities in the represented constellation space is proportional to their meaning dissimilarity. For example, if we adopt the energy-base semantic distance to measure the meaning dissimilarity between two entities, the edge server will train a projection (encoding) function by minimizing the following loss function:
\begin{eqnarray}
L &=&\! \sum\limits_{\scriptstyle \langle \tbe,\tbr,\tbe' \rangle\in {\cK}, \atop \scriptstyle \langle \tbe,\tbr',\tbe'' \rangle\in {\cK}^-} \!\max\{0, \sigma + {\|\tbe+\tbr-\tbe'\|^2}  \nonumber \\
&&\;\;\;\; - \|\tbe_t+\tbr'-\tbe''\|^2\} ,
\label{eq_TransELoss}
\end{eqnarray}
where ${\cK}$ and ${\cK}^-$ correspond to the sets of valid and invalid entity-relation connecting triplets in the edge server's local knowledge base, respectively. $\sigma$ is the average distance between the valid (positive) and invalid (negative) entity-relation connecting triplets. The trained projection function, denoted by $\Theta(\cdot)$, will be sent to the source user to encode the explicit semantics into a sequence of constellation representation of symbols before sending to the physical channel.  In this way, the entities representing similar meanings will have relatively short Euclidean distance, compared to those with different meanings. When the encoded signal has been corrupted by the channel fading or noise, each symbol (entity) will be more likely to be misrepresented by the semantic similar symbol instead of semantic dissimilar ones. In other words, our proposed encoding strategy is more robust against semantic misinterpretation than existing semantic-irrelevant communication solutions, especially for applications that can tolerate a certain  symbol-level loss during the semantic interpretation.

Compared to many existing semantic communication solutions which directly encode the labels or features of objects into a sequence of binary bits to be sent based on the traditional channel coding solution, our proposed semantic-aware encoding solution can encode more semantic symbols in each constellation by separating different symbols by their meaning differences, and therefore offers much improved communication efficiency and enhanced robustness against the channel corruption, especially when being combined with our proposed semantic decoder which will be discussed later in this section.

\begin{figure}
\centering
    \begin{minipage}{.45\textwidth}
    \centering
    \includegraphics[width=\textwidth]{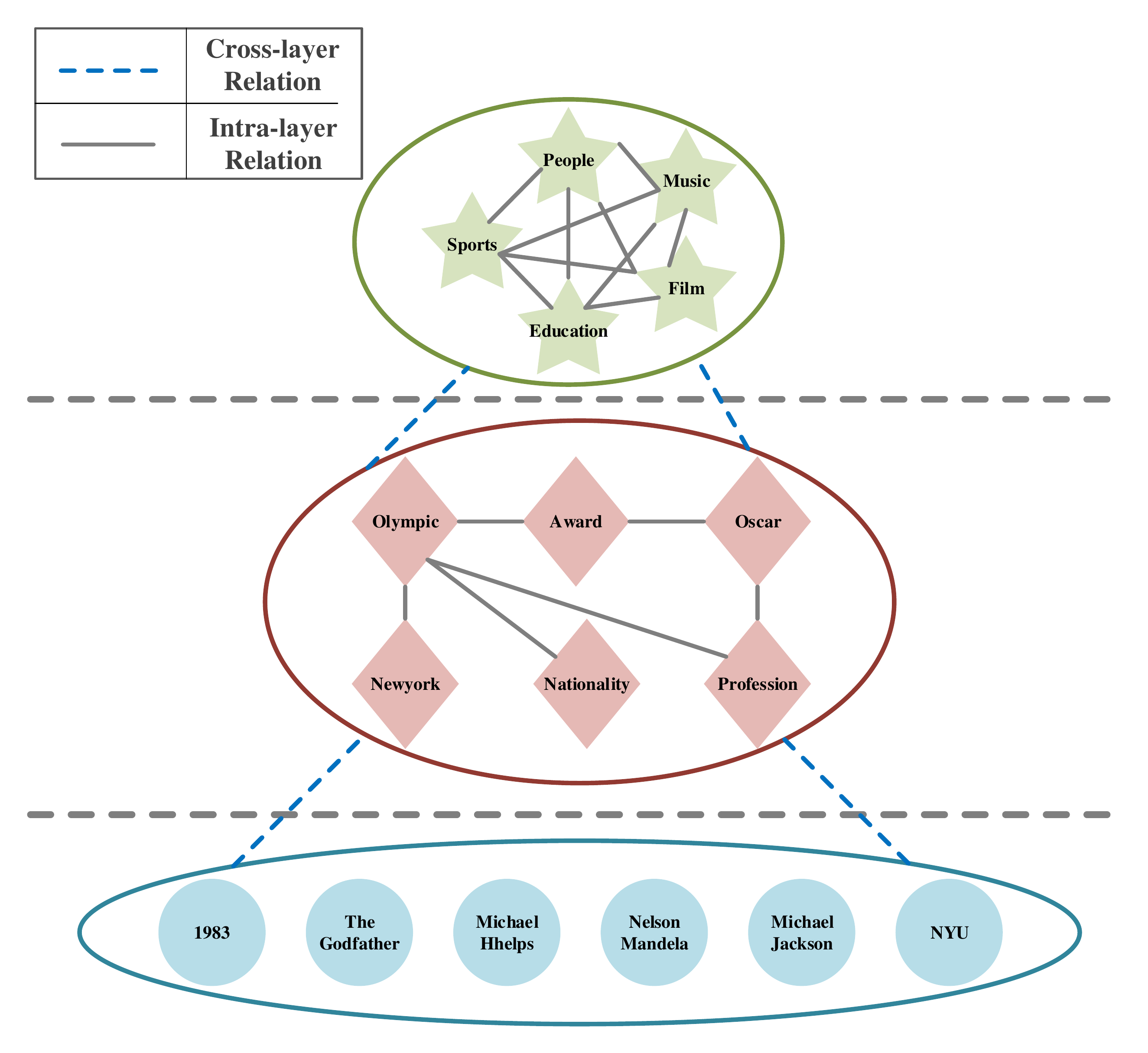}
    \subcaption{}
    \label{Fig_constellations_TieredSemantics}
    \end{minipage}
    \begin{minipage}{.45\textwidth}
        \centering
    \begin{minipage}{.45\textwidth}
        \centering
    \includegraphics[width=\textwidth]{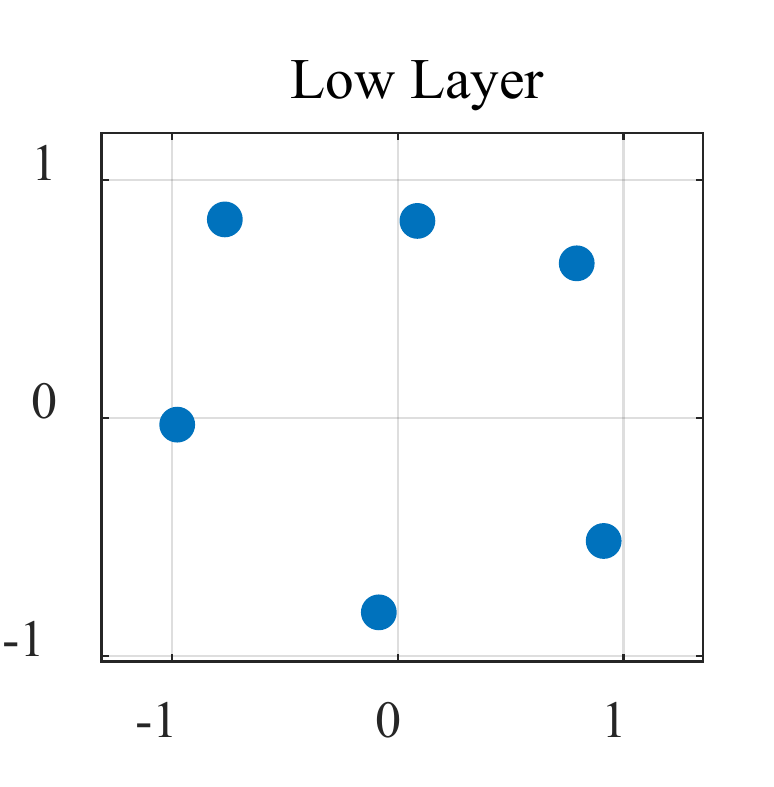}
    \subcaption{}
    \label{Fig_constellations_Low}
    \end{minipage}
    \begin{minipage}{.45\textwidth}
    \includegraphics[width=\textwidth]{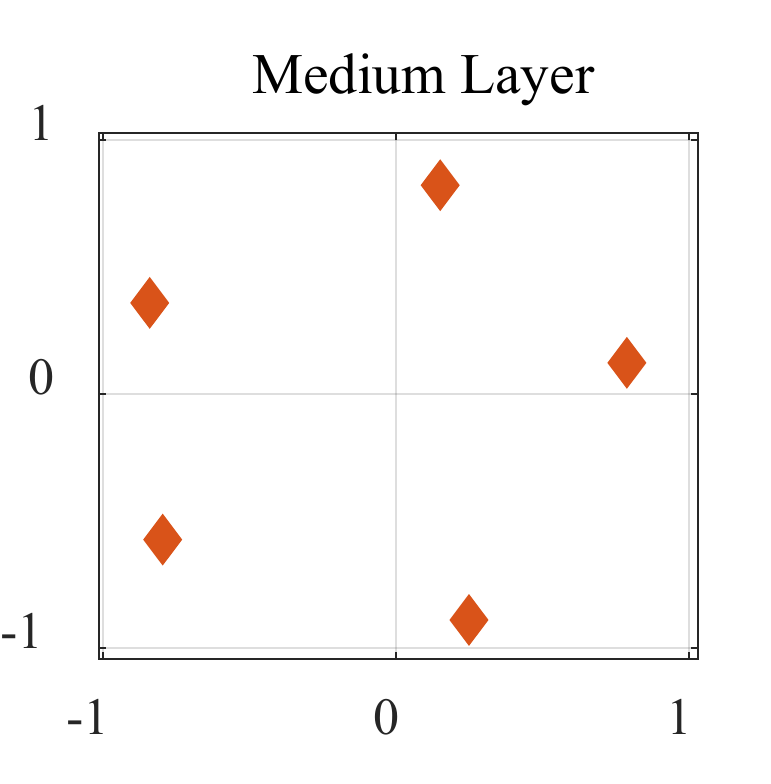}
    \subcaption{}
    \label{Fig_constellations_Medium}
    \end{minipage}

    \begin{minipage}{\textwidth}
    \centering
    \begin{minipage}{.45\textwidth}
        \centering
    \includegraphics[width=\textwidth]{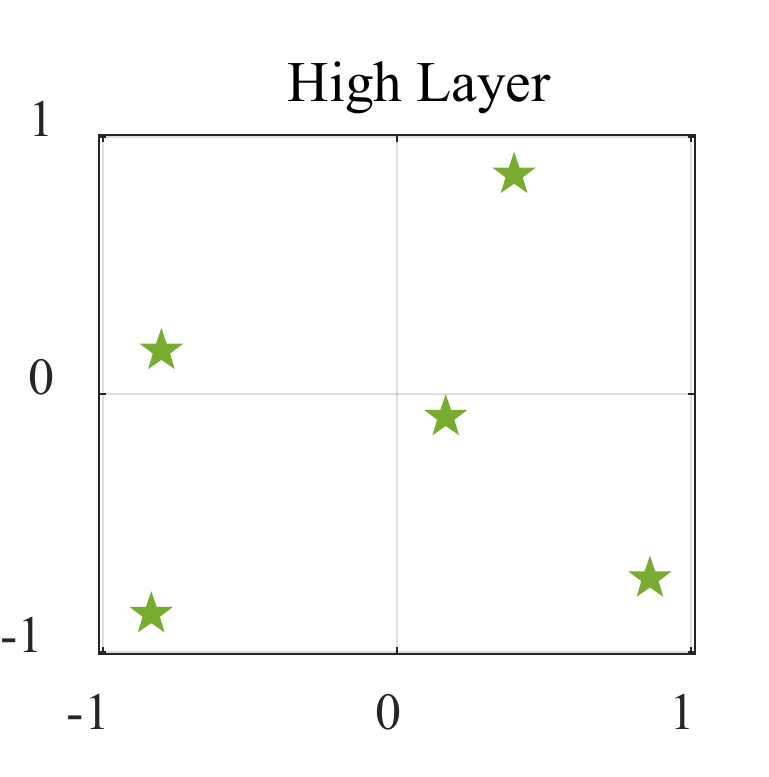}
    \subcaption{}
    \label{Fig_constellations_High}
    \end{minipage}
    \begin{minipage}{.45\textwidth}
        \centering
    \includegraphics[width=\textwidth]{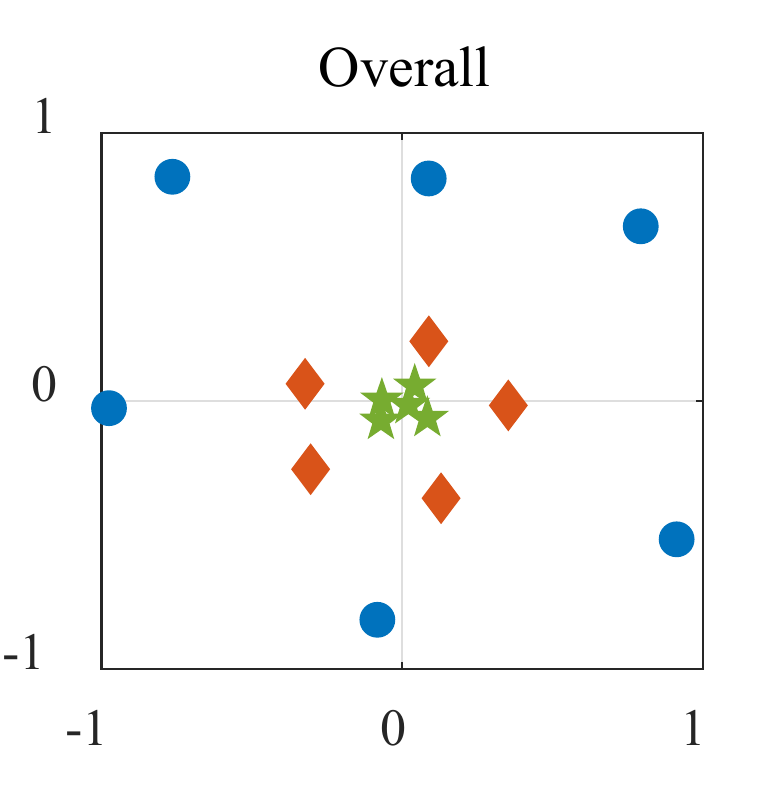}
    \subcaption{}
    \label{Fig_constellations_Overall}
    \end{minipage}
    \end{minipage}
    \end{minipage}
\caption{\footnotesize{(a) A multi-layer representation of semantics of a message consisting of 17 knowledge entities across three abstraction layers and (b)-(d) their corresponding semantic constellations represented with entities in three individual layers and (e) all the entities represented in a single semantic constellation space.}}
\label{Fig_constellations}
\end{figure}

In Fig. \ref{Fig_constellations}, we consider an example message consisting of semantic meaning represented by 17 knowledge entities generated from FB15K-237 knowledge dataset divided into three abstraction layers. It is known that each entity in FB15K-237 has over 200 attributes and therefore to transmit all these explicit semantics will require sending 3,400 real valued data coded into binary bits when applying the traditional communication solution. In our proposed solution, the multi-layer representation of semantics will be encoded into a 2-dimensional semantic constellation with real and image parts. We can either encode explicit semantics of different layers into separate constellation representations as shown in Figs. \ref{Fig_constellations}(b)-(d) or encode all the semantic entities into a single semantic constellation as shown in Fig. \ref{Fig_constellations}(e). As mentioned earlier, entities in the higher layers generally have more directly connected relations with each other and therefore their representations in the constellation space will be more densely located. Also, entities in the lower layers tend to have less directly connected relations and therefore are located more separately. 




\subsubsection{Training Semantic Decoder at the Edge Server}
The destination user will also rely on the edge server to train a semantic decoder, denoted by $\Theta^{-1}(\cdot)$, to recover the explicit semantics from the received signal. In particular, suppose the source user sends an encoded signal $\hat \bv^E$. The signal received by the destination user is given by
\begin{eqnarray}
\hat \bw^E = g \hat \bv^E + \delta,
\end{eqnarray}
where $g$ is the channel fading coefficient and $\delta$ is the additive Gaussian noise.

During the training phase, the destination user will directly upload the received signal $\hat \bw^E$ to the edge server. The edge server can then decide whether or not to fully decode the explicit semantics before sending to the semantic interpreter. In particular, we consider two semantic decoding schemes for the destination user:
\begin{itemize}
    \item {\em Hard Decoding:} the destination user first recovers the explicit semantics sent by the source user from its received signal and then sends the recovered explicit semantics to the semantic interpreter for inferring the implicit semantics. Since the semantic encoding function used by the source user can also be trained by the edge server, during the training phrase, when the edge server receives the noisy version of semantic signal uploaded from the destination user, it can estimate the  fading coefficient and additive noise level of the channel between the source and destination users and then calculate a decoding function that maps a noisy version of the low-dimensional semantic representation of signals received by the destination user to the high-dimensional graphical form of explicit semantics sent from the source user.  
    \item {\em Soft Decoding:} the noisy version of the received signal will be directly sent to the semantic interpreter for generating the reasoning paths. 
    Suppose the energy-based semantic encoding function trained by (\ref{eq_TransELoss}) has been used. The semantic interpreter, once received a noisy version of explicit semantic representation $\hat \bw^E$, 
    will sequentially extend a reasoning path with length $J$ ($J$ relations being decided) from $\hat \bw^E$ as follows: 
    \begin{eqnarray}
    \hat \bp = \sum\limits^J_{j=1} \tbr_j + \hat \bw^E.
    \end{eqnarray}

    \blu{Similarly, the semantic decoding function can also correspond to a neural network model which generates semantic reasoning paths based on the noisy version of explicit semantics by following a learned policy model $\bpi_{E}$. We will give a more detailed discussion in the next subsection. }

\end{itemize}

The detailed semantic encoding and decoding processes are illustrated in Algorithm \ref{Algorithm_SemanticCoding}.

\begin{algorithm}
\caption{Training Semantic Encoder and Decoder}
\label{Algorithm_SemanticCoding}
\footnotesize
\tcp*[h]{Training Phase}\\
{\bf Input:} Entities and relations of previously observed expert paths\\
{\bf Output:} The project function $\Theta$ \\
\For{each server $k \in [K]$}{
{Initialize $\Theta$}\\
\While{$L \geq Threshold$}
{Update $\Theta$ w.r.t $\sum\limits_{\scriptstyle \langle \tbe,\tbr,\tbe' \rangle\in {\cK}, \atop \scriptstyle \langle \tbe,\tbr',\tbe'' \rangle\in {\cK}^-} \nabla \{\sigma + {\|\tbe+\tbr-\tbe'\|^2}- \|\tbe_t+\tbr'-\tbe''\|^2\}$}
}
{\bf Return} $\Theta$\\
\tcp*[h]{Encoding Phase}\\
{\bf Input:} Explicit semantics identified by the semantic identifier\\
{\bf Output:} Low-dimensional representations of semantics
\begin{itemize}
    \item Source user receives $\Theta$
    \item $\hat{\bv}^E \leftarrow \Theta(\bv^E)$
\end{itemize}
\tcp*[h]{Decoding Phase}\\
{\bf Input:} $\hat{\bw}^E$ received by the destination user\\
{\bf Output:} A set of reasoning paths that best interpret the implicit semantics \\
\emph{Destination user uploads $\hat{\bw}^E$ to edge server}\\
\eIf{hard decoding}{
    $\bw^E \leftarrow \Theta^{-1}\left(\hat{\bw}^E\right)$}{
    $\hat{\bp} \leftarrow \sum\limits^J_{j=1} \tbr_j + \hat \bw^E$}
\end{algorithm}

\subsubsection{Training Semantic Interpreter at the Edge Server}
To fully recover the semantic meaning of the message, the edge server will need to train a semantic reasoning policy that can generate a set of possible reasoning paths based on the observed explicit semantics. We propose a generative adversarial imitation learning-based solution for the source user to assist the training of the semantic interpreter based on its previously observed expert paths. In this approach, the source user and semantic interpreter will compete with each other in training a policy to generate reasoning paths that are indifferentiable to the expert paths observed by the source user.  More specifically, during the training phrase, the source user will randomly sample some initial entities in the expert paths as explicit semantics which will be sent to the edge server. The edge server will then try to generate a set of reasoning paths with length $J$ extended from the received explicit semantics to be sent back to the source user.
The source user will compare the occupancy measure of the received reasoning path with that of the expert paths to calculate the semantic distance to be sent back to the edge server. The edge server can adopt the standard SGD-based approach to iteratively learn a policy to minimize the semantic distance between its generated paths and the expert paths as the above process repeats, i.e., the edge server will try to minimize the following problem,
\begin{eqnarray}
\begin{aligned}
\min\limits_{\bpi_D}  \bigg( -H \left( \bpi_D \right) + \lambda &\Big( \mathbb{E}_{\bc_{\bpi_E}\sim \Delta^E}  \left[\varpi^*_{\phi} \left(\bc_{\bpi_E} \left( \bq \right)\right)\right]  \\
& + \mathbb{E}_{\bc_{\bpi_D} \sim \Delta^D}\left[ \varpi^*_{\phi} \left( \bc_{\bpi_D} \left( \bp \right) \right) \right]\Big)\bigg),
\label{eq_ProblemInterpreter}
\end{aligned}
\end{eqnarray}
where $\varpi^*_\phi$ is the semantic evaluator that is learned by the source user based on the observed expert  reasoning paths which will be discussed in the next subsection.

At the beginning of the training phrase, the semantic interpreter can be a randomly initiated graph convolutional network (GCN) to generate random paths from any given explicit semantics based on the local knowledge base associated with the destination user. As will be proved later, the semantic interpreter associated with a single source user will always converge to a stationary reasoning policy $\bpi_D$.
From (\ref{eq_ProblemInterpreter}), we can observe that $\varpi^*_{\phi}$ is in fact a discriminator that tries to differentiate the paths generated by the semantic interpreter from  the expert paths observed by the source user. 




\subsubsection{Training Semantic Evaluator at the User}
As mentioned earlier, the source user will not reveal its expert paths to the edge servers of the destination users but will assist the training process of the reasoning mechanism by reporting the difference (semantic distance) between the  semantic reasoning paths interpreted by the edge servers and the expert paths. For example, the source user can train a semantic evaluator, a neural network to differentiate the interpreted meaning and the meaning of the expert paths, i.e., the main objective of the semantic evaluator is to maximize the cross entropy between the occupancy measure of the expert paths and that of the interpreted paths,

\begin{eqnarray}
\bvarpi^*_\phi \;\;\;\;= && \arg \max\limits_{\bvarpi_\phi} \bigg( - H \left( \bpi_D \right) \nonumber\\
&& + \lambda \Big( \mathbb{E}_{\bc_{\bpi_E}\sim \Delta^E}  \left[\bvarpi_{\phi} \left(\bc_{\bpi_E} \left( \bq \right)\right)\right] \nonumber \\
&& + \mathbb{E}_{\bc_{\bpi_D} \sim \Delta^D}\left[ \bvarpi_{\phi} \left( \bc_{\bpi_D} \left( \bp \right) \right)\right] \Big) \bigg).
\label{eq_SemanticEvaluator}
\end{eqnarray}

\begin{theorem}
For any given $\bpi_D$, there always exists an optimal solution for problem (\ref{eq_SemanticEvaluator}) under a given $\bpi_D$. As long as $\bpi_D$ converges to the optimal solution of problem (\ref{eq_ProblemInterpreter}), then the resulting semantic distance between the expert paths and the reasoning paths generated by the semantic interpreter approaches zero.
\label{Theorem_GAN}
\end{theorem}
\begin{IEEEproof}
See Appendix \ref{Proof_GAN}.
\end{IEEEproof}



\subsection{iRML with Multi-Tier Collaborative Reasoning}
\label{Subsection_iRMLCollaberativeReasoning}

Let us consider a multi-tier cloud/edge network with decentralized knowledge bases. In this case, each edge server or CDC can only access a subset of knowledge concepts and relations. 
Motivated by the fact that knowledge bases of edge severs within the same tier may exhibit similar reasoning habits when reasoning about the relations and reasoning paths among entities, we propose a federated GCN-based model aggregation solution to jointly construct the reasoning mechanism at the semantic interpreters among the same-tier edge severs. We assume there exists a coordinator that can communicate with all the servers within the same tier to periodically coordinate their local model training process. The coordinator can either be a higher-tier server or one of the same-tier server selected to maintain and update the global model. 
Semantic encoder and decoder will be locally trained and updated by each edge server to be broadcast to the local source and destination users in its coverage area.

Two key challenges of the collaborative model training across decentralized knowledge bases are: (1) {\em cross-layer (knowledge base) reasoning}: a single message received by a source user may involve semantic entities stored at multiple layers of edge servers and/or CDC; and (2) {\em missing (inter-knowledge-base) relations}: different from the sample-based dataset, in this paper, we consider the relation-based semantic reasoning in which there may exist  relations between entities at different knowledge bases stored at different edge servers. Unfortunately, these inter knowledge base relations are discarded due to the decentralized distribution of knowledge bases.

To address the first challenge, we introduce an index list at each edge server to specify all the other knowledge entities with the associated edge servers. Note that this index only consists of links between knowledge entity labels and their located edge servers. No relations nor attributes of the entities will be stored or revealed in this index. To seek interpretation of semantics, each user will submit its explicit semantics to the closest edge server, which will then distribute the entities in the received explicit semantics into the corresponding edge servers according to the index. Each edge server will then generate a reasoning path from the received entities based on its local knowledge base and the learned reasoning policy. If the generated paths only involve entities in the local knowledge base of the edge server, these paths will be sent back to the local edge server of the user. We refer to the reasoning policy that  decides the relations connecting entities within the same layer of knowledge bases as the {\em intra-server policy}. There are cases some reasoning paths generated by an edge server may need to extend to entities at other same-tier or even cross-tier edge servers' knowledge bases, which we referred to as the {\em intra-layer policy} or {\em cross-layer policy}. 


In this case, the source user will need to also guide each edge server to train a {\em cross-server policy} which  decides the next entities in another edge server based on the currently observed or inferred entities in the current edge server. Since  no knowledge base has the inter-knowledge-base relations, it is generally impossible to know the detailed types of relations that exist between entities cross different knowledge bases. The cross-server policy can therefore only decide a binary  relation (1 or 0, exist or no relation) specifying whether or not there exist a relation between an entity in other knowledge bases and the current one. 

Both intra-server policy and cross-server policy can be collaboratively trained by multiple edge servers. Let us now describe the detailed training and model coordination process as follows.


\begin{itemize}
    \item[(1)] Intra-server Policy Coordination: 
    Each edge server will perform local model training for the semantic interpreter following the same procedures described in Section \ref{Subsection_iRMLSingleEdgeServer}. The locally trained model parameters of the semantic interpreters at edge servers can be periodically coordinated during the training phrase. Here, we will mainly discuss the coordination procedures between edge servers via a coordinator.
    At the beginning of the training phase, the coordinator will broadcast a global model to every edge server. For example, if we adopt a commonly used setup and assume the semantic interpreter of each edge server is a GCN with two-layer fully-connected convolutional layers, we have
\begin{eqnarray}
\bpi_{D_k} = \sigma (\mathbf{w}^2_k, \tau(\mathbf{w}^1_k, \bp^E_k)),
\end{eqnarray}
where we use subscript $k$ to denote the parameters associated with the $k$th edge server, $\sigma(\cdot)$ and $\tau(\cdot)$ are activation functions which can be ReLU, softmax, and sigmod. Let $\mathbf{w}_k$ be the concatenation of both layers of model parameters, i.e., we have $\mathbf{w}_k = \langle \mathbf{w}^1_k, \mathbf{w}^2_k \rangle$. Recall the local loss function of the semantic interpreter in (\ref{eq_ProblemInterpreter}) is given by $F_k = -H(\bpi_{D_k})+\lambda \Gamma (\bpi_{E_k}, \bpi_{D_k})$. Suppose edge servers upload their local models every $E$ rounds of local SGD iterations. 
We adopt the FedAvg\cite{mcmahan2017communication}, one of the most commonly used federated learning algorithms for model coordination, to coordinate the model training among edge severs. We can therefore write the model aggregation among $K$ collaborative edge servers 
as $\mathbf{\bar w} = \sum^K_{k=1} \gamma_k \mathbf{w}_k$ where $\gamma_k=\frac{n_k}{\sum^K_{k=1} n_k}$ and $n_k$ is the number of entities in the knowledge base of edge server $k$. 

    \item[(2)] Cross-server Policy Coordination: each edge server can also train a cross-server semantic reasoning policy that only specifies a binary relation connecting each of its entities with any other entities in the knowledge bases of other edge servers. The model training and aggregation processes are exactly the same as the intra-server policy case with a different action space, which consists only of  entities from other edge servers.

\end{itemize}

The detailed algorithm is presented in Algorithm \ref{Algorithm_Collaborative}.

\begin{algorithm}
\footnotesize
\caption{Collaborative Reasoning Mechanism}
\label{Algorithm_Collaborative}
{\bf Input:} Expert path set $\mathbf{\pi}_E$, encoded signal ${\bv}^E$ sent by source user  \\
{\bf Output:} Learned policy $\mathbf{\pi}^*_D$\\
Initialize ${\mathbf{w}^{\left(1\right)}}_k = {\bar{\mathbf{w}}}^{\left(1\right)}$ and ${\mathbf{\pi}}_{D_k}$\\
Randomly generate a path ${p_k}^{(1)}$ according to $\mathbf{\pi}_{D_k}$ and ${\bv}^E_k$ \\
\For{$t=1,\cdots,T$}{
    \eIf{$t+1 \notin [nE], n = 0, 1, 2, \cdots$}{
    \For{each server $k \in [K]$}{
    Generate ${p_k}^{(t)}$ according to $\mathbf{\pi}_{D_k}^{(t)}$\\
    Send ${p_k}^{(t)}$ back to source users;\\
    Get semantic difference\\
    ${\boldsymbol{g}^{\left(t\right)}}_{\mathbf{w}_k} \leftarrow \nabla_{\mathbf{w}_k}F_k\left({\mathbf{w}^{\left(t\right)}}_k\right)$ \\
   $\mathbf{w}_{k}^{(t+1)}=\mathbf{w}_{k}^{(t)}-\eta_{t} \boldsymbol{g}_{\mathbf{w}_{k}}^{(t)}$;
    \\\tcp*[h]{Update local parameters.}\\
    Update $\mathbf{\pi}_{D_k}^{(t+1)}$;}}
    {$\bar{\mathbf{w}}^{(t+1)} \leftarrow \sum_{k=1}^{K} \gamma_k{\mathbf{w}}^{(t+1)}_{k}$;    \\\tcp*[h]{Aggregation at the coordinator.}\\
    The coordinator broadcasts $\bar{\mathbf{w}}^{(t+1)}$ back to edge servers;\\
    Update $\mathbf{\pi}_{D_k}^{(t+1)}$;}}
\end{algorithm}






Many existing works in federated learning have already verified that the FedAvg algorithm is able to converge to a global optimal model when training with a global knowledge base consisting of all the knowledge information of the collaborative edge servers. Unfortunately, since the division of knowledge entities between decentralized knowledge bases results in the loss of information about the inter-knowledge-base relations, the collaborative model training in our case cannot converge to the global optimal solution. However, in many real-world scenarios, the number of inter-knowledge-base relations can be very limited, especially compared to the relations connecting entities within each knowledge base. For example, in the multi-tier computing network, the regionally-shared knowledge entities are generally more closely related to each other than other entities from other regionally-shared knowledge base. In this case, the model trained by our proposed collaborative training solution will approach to the close neighborhood of the global optimal solution.

Let us now derive the theoretical convergence bound to capture the above-mentioned performance gap between the collaboratively trained  model based on decentralized knowledge bases and the global optimal model trained with the complete global knowledge information.

Consider $K$ same-tier edge servers collaborate to train a shared semantic interpretation  model. Let $N$ be the total number of knowledge entities of a global knowledge base, a combination of knowledge bases of all $K$ collaborative edge servers. We use $\mathbf{A}$ and $\mathbf{X}$ to denote the adjacent matrix and feature vector of  the possible relations between entities in the global knowledge base. Let $\mathbf{A}_k$ and $\mathbf{X}_k$ be the adjacent matrix and feature vector of the relations between knowledge entities in edge server $k$. We follow a commonly adopted setting and quantify the heterogeneity of local knowledge data distribution across collaborative edge servers as $\rho=F^{*}-\sum_{k=1}^{K} \gamma_{k}F_{k}^{*}$.

Before introduce the main theoretical result, let us introduce the following assumptions.

\begin{assumption}\label{assumption_continous}
The local function $F_1,...,F_k$ are all $L$-smooth, i.e., there exists a constant $L>0$, such that $F_{k}\left(\mathbf{w}_{i}\right) \leq F_{k}\left(\mathbf{w}_{j}\right)+\left(\mathbf{w}_{i}-\mathbf{w}_{j}\right)^{T} \nabla F_{k}\left(\mathbf{w}_{j}\right)+\frac{L}{2}\left\|\mathbf{w}_{i}-\mathbf{w}_{j}\right\|_{2}^{2}, \forall k \in \mathcal{K}$.
\end{assumption}

\begin{assumption}\label{assumption_convex}
The local function $F_1,...,F_k$ are all $\mu$-strongly convex, i.e., there exists a constant $\mu>0$, such that $F_{k}\left(\mathbf{w}_{i}\right) \geq F_{k}\left(\mathbf{w}_{j}\right)+\left(\mathbf{w}_{i}-\mathbf{w}_{j}\right)^{T} \nabla F_{k}\left(\mathbf{w}_{j}\right)+\frac{\mu}{2}\left\|\mathbf{w}_{i}-\mathbf{w}_{j}\right\|_{2}^{2}, \forall k \in \mathcal{K}$.
\end{assumption}

\begin{assumption}\label{assumption_BoundedGradient}
The expected squared norm of local gradients is uniformly bounded, i.e., $\mathbb{E}\left\|\nabla F_{k}\left(\mathbf{w}_{t}^{k}\right)\right\|^{2} \leq \sigma_{L}^{2}, \forall k \in \mathcal{K}$ and $t = 1,\cdots,T-1$.
\end{assumption}

Assumptions \ref{assumption_continous}-\ref{assumption_BoundedGradient} are commonly adopted settings in most federated learning solutions and have already been verified to be satisfied in many practical scenarios\cite{fedavg,mills2019communication,konevcny2016federated}. We can prove the following results.

\begin{theorem}\label{Theorem_ConvergeFL}
Suppose Assumptions \ref{assumption_continous}-\ref{assumption_BoundedGradient} hold and the learning rate is given by $\eta_{t}=\frac{2}{\mu} \frac{1}{\zeta+t}$, where $\zeta=\max \left\{8 \kappa, E\right\}$ and $\kappa=\frac{L}{\mu}$. The  coordinated model parameters $\bar{w}_T$ trained by $K$  collaborative edge servers after $T$ iterations of local SGD training satisfy
\begin{equation}
\mathbb{E}\left[F_k\left(\mathbf{\bar{{w}}}_{T}\right)\right]-F^{*} \leq \frac{2 \kappa}{\zeta+T-1}\left(\frac{\Omega}{\mu}+\frac{2 L_{p}}{\mu N}\mathcal{D}\right),
\end{equation}
where $L_p$ is a constant satisfying Lipstchitz continuity and $\Omega$ and $\mathcal{D}$ are given by 

\begin{equation}
\Omega=4\left(1+2(E-1)^{2}\right) \sigma_{L}^{2}+4 L \rho+\frac{\mu^{2} \zeta}{4}\left\|\mathbf{w}_{1}-\mathbf{w}^{*}\right\|^{2}
\end{equation}
and
\begin{equation}
    \mathcal{D}=\left\|K \mathbf{X}_{k}^{T} \mathbf{A}_{k}^{T} \mathbf{A}_{k}^{T} \mathbf{A}_{k} \mathbf{A}_{k} \mathbf{X}_{k}-\mathbf{X}^{T} \mathbf{A}^{T} \mathbf{A}^{T} \mathbf{A} \mathbf{A} \mathbf{X}\right\|^{2}.
\end{equation}
\end{theorem}

\begin{IEEEproof}
See Appendix \ref{Proof_ConvergeFL}.
\end{IEEEproof}

From Theorem \ref{Theorem_ConvergeFL}, we can observe that the convergence bound of our proposed collaborative model training is mainly affected by  two key parameters: $\Omega$ and $\mathcal{D}$. The value of $\Omega$ is mainly affected by the total number of local SGD iterations and the  heterogeneity of knowledge information across collaborative edge servers. The value of $\mathcal{D}$ quantifies the divergence between the local and global adjacent matrices which directly reflects the performance gap between the collaboratively trained  model based on decentralized knowledge bases and the global model trained with the global knowledge base.

%




\section{Experimental Result}
\label{Section_NumericalResult}

\subsection{Dataset and Simulation Setup}
In this section, we evaluate the semantic communication performance based on iRML. To demonstrate the generality of iRML, 
we simulate the semantics of messages generated by three popular real-world knowledge datasets, Cora, Citeseer, and FB15K-237, composed of different types of knowledge entities. In particular, Cora and Citeseer are  citation datasets, consisting of 2708 and 3327 knowledge entities (scientific papers), 5429 and 4732 relations, and 1433 and 3707 features, respectively. The semantics sampled from these two datasets can be used to simulate the meaning represented by a sequence of topics, studied in the scientific articles. FB15K-237 dataset is a dictionary-based human knowledge dataset consisting of 14,541 entities and 237 types of relations. The semantics sampled from FB15K-237 dataset may be used to simulate the meaning of messages composed of the sampled words connected with relations. The set of expert semantic reasoning paths observed by each source user is obtained by random sampling a set of paths based on a two-sided breadth first path searching algorithm. Since this path sampling algorithm always searches for the shortest paths connecting any given pair of entities, the establish set of expert paths can simulate the semantic reasoning mechanism of the users with more straight-forward inference preference.  
The semantics of each individual message is simulated by a path with the first entity considered as the explicit semantics and the rest of the relations and entities are considered as implicit semantics.      


We consider the energy-based semantic distance and set semantic interpreter as a two-layer fully-connected graph convolutional networks (GCNs) with the output being normalized by a softmax function. The semantic evaluator consists of a hidden layer and an output layer also being normalized by a sigmoid function.  

    \begin{figure}[htbp]
    \centering
    \begin{minipage}{.45\linewidth}
    \centering
    \includegraphics[width=\textwidth]{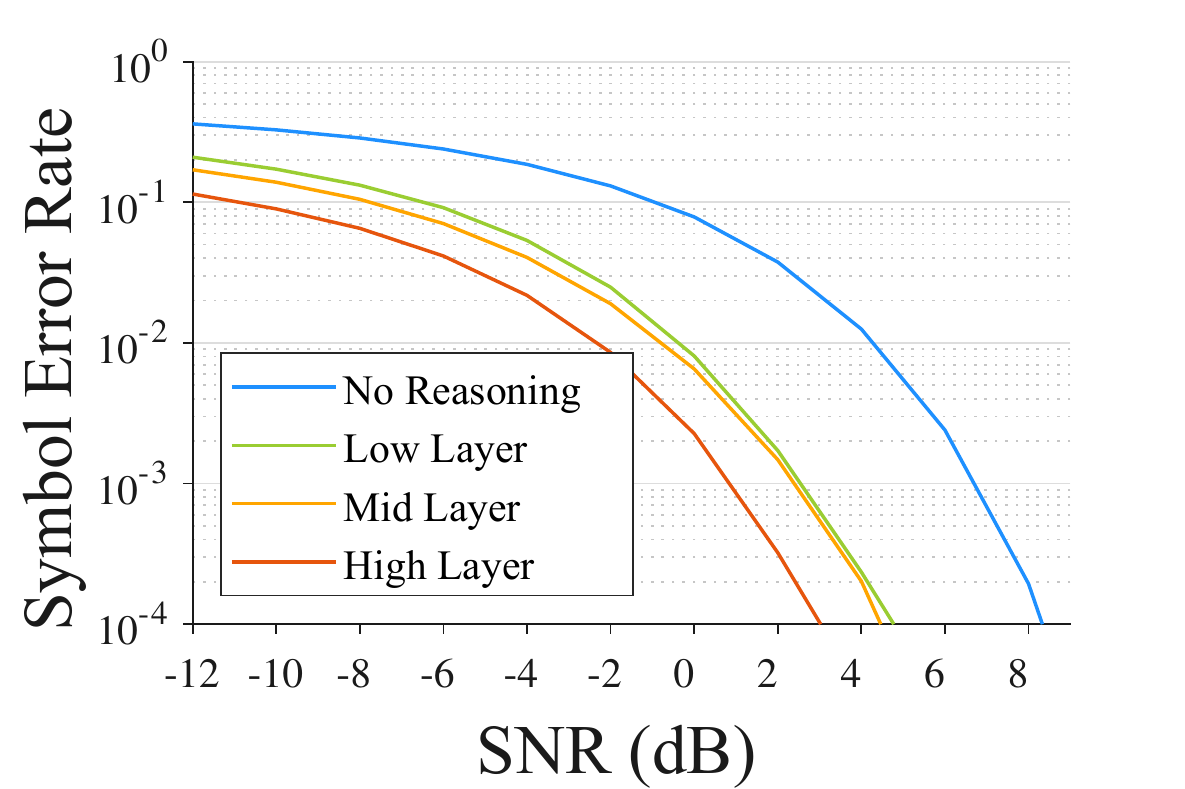}
    \captionsetup{labelformat=empty}
    \caption*{\footnotesize{(a)}}
    \label{Fig_SymbolicErrorRate_hard}
    \end{minipage}
    \begin{minipage}{.45\linewidth}
    \centering
    \includegraphics[width=\textwidth]{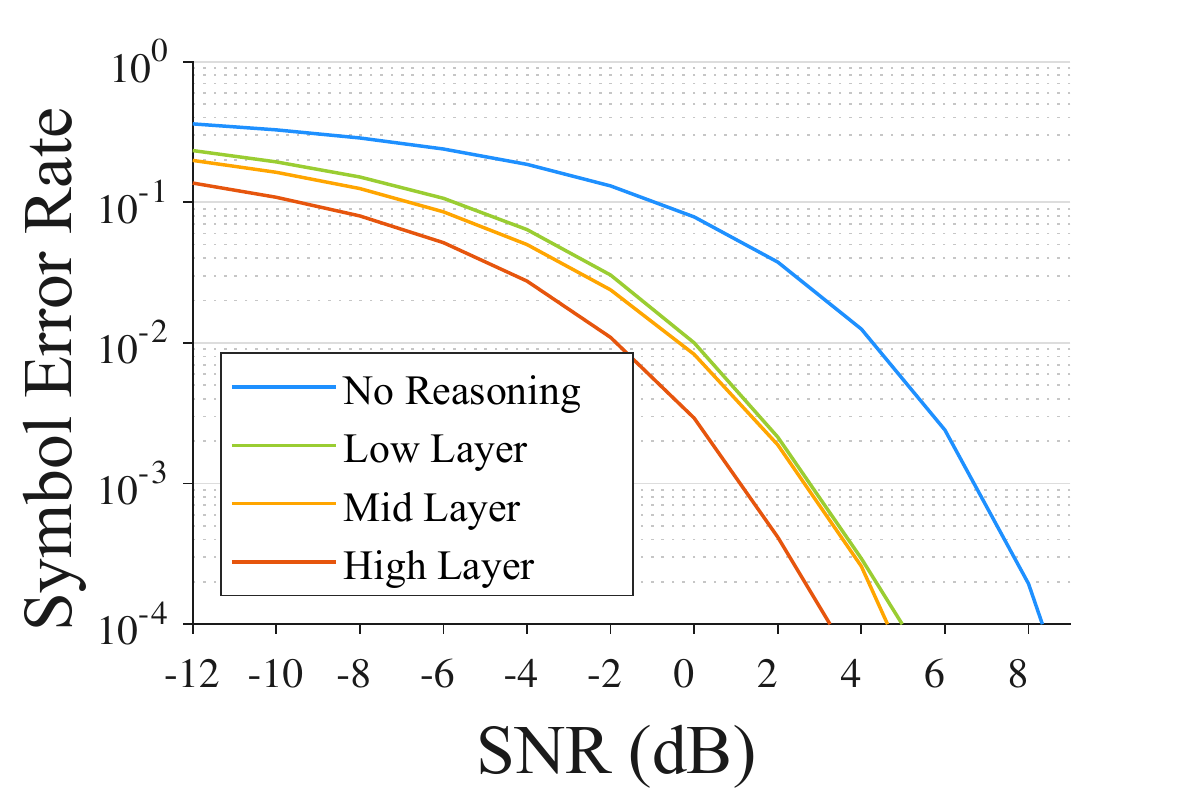}
    \captionsetup{labelformat=empty}
    \caption*{\footnotesize{(b)}}
    \label{Fig_SymbolicErrorRate_soft}
    \end{minipage}
\caption{\footnotesize{Symbol error rate of semantic symbols (entities) associated with different abstraction layers when being decoded with (a) hard decoding and (b) soft decoding under different SNRs.}}
    \label{Fig_SymbolicErrorRate}
\end{figure}

 \begin{figure}[htbp]
    \centering
    \begin{minipage}{.45\linewidth}
    \centering
    \includegraphics[width=\textwidth]{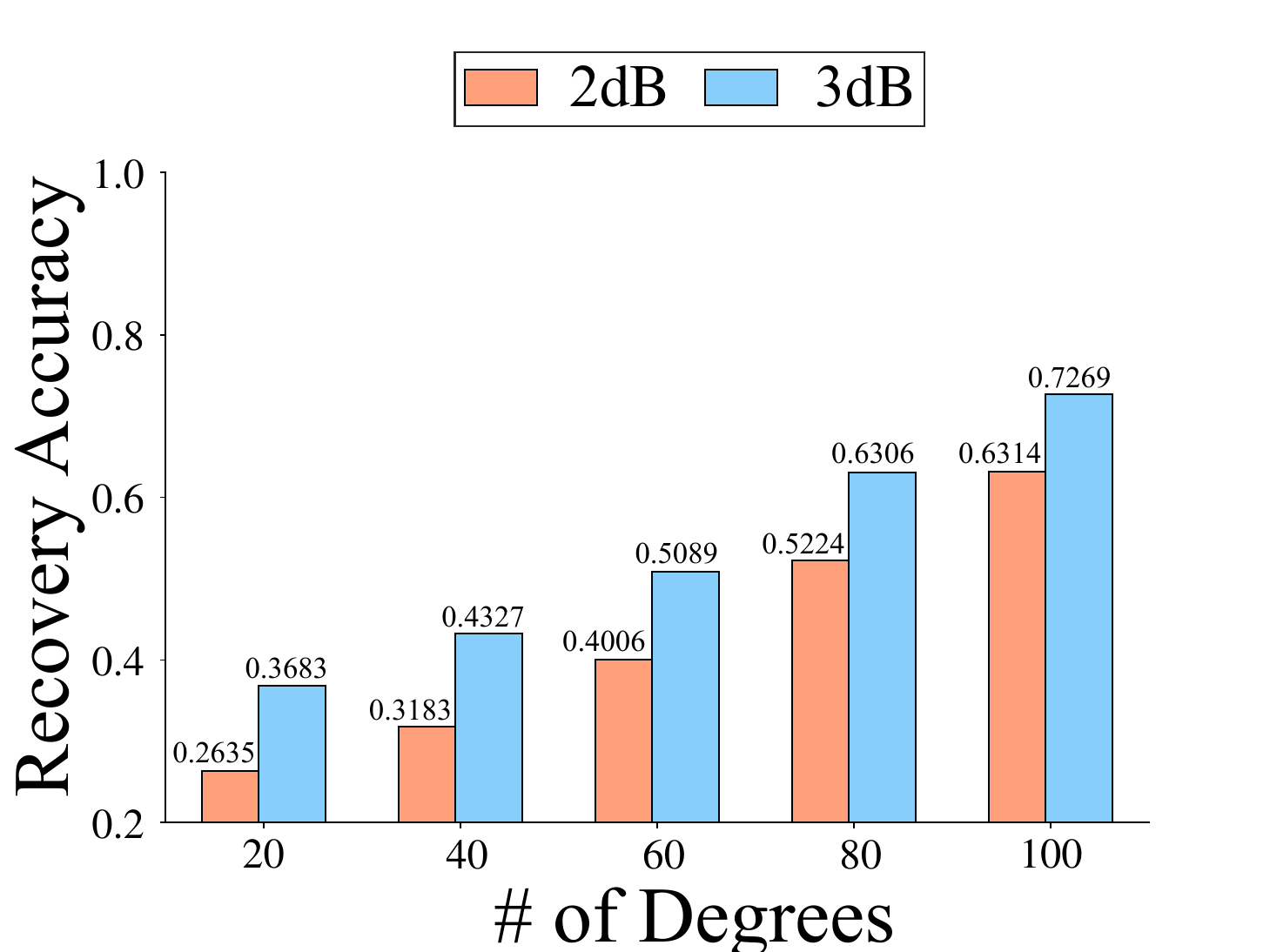}
    \captionsetup{labelformat=empty}
    \caption*{\footnotesize{(a)}}
    \label{Fig_ACCvsNumDegree_lowSNR}
    \end{minipage}
    \begin{minipage}{.45\linewidth}
    \centering
    \includegraphics[width=\textwidth]{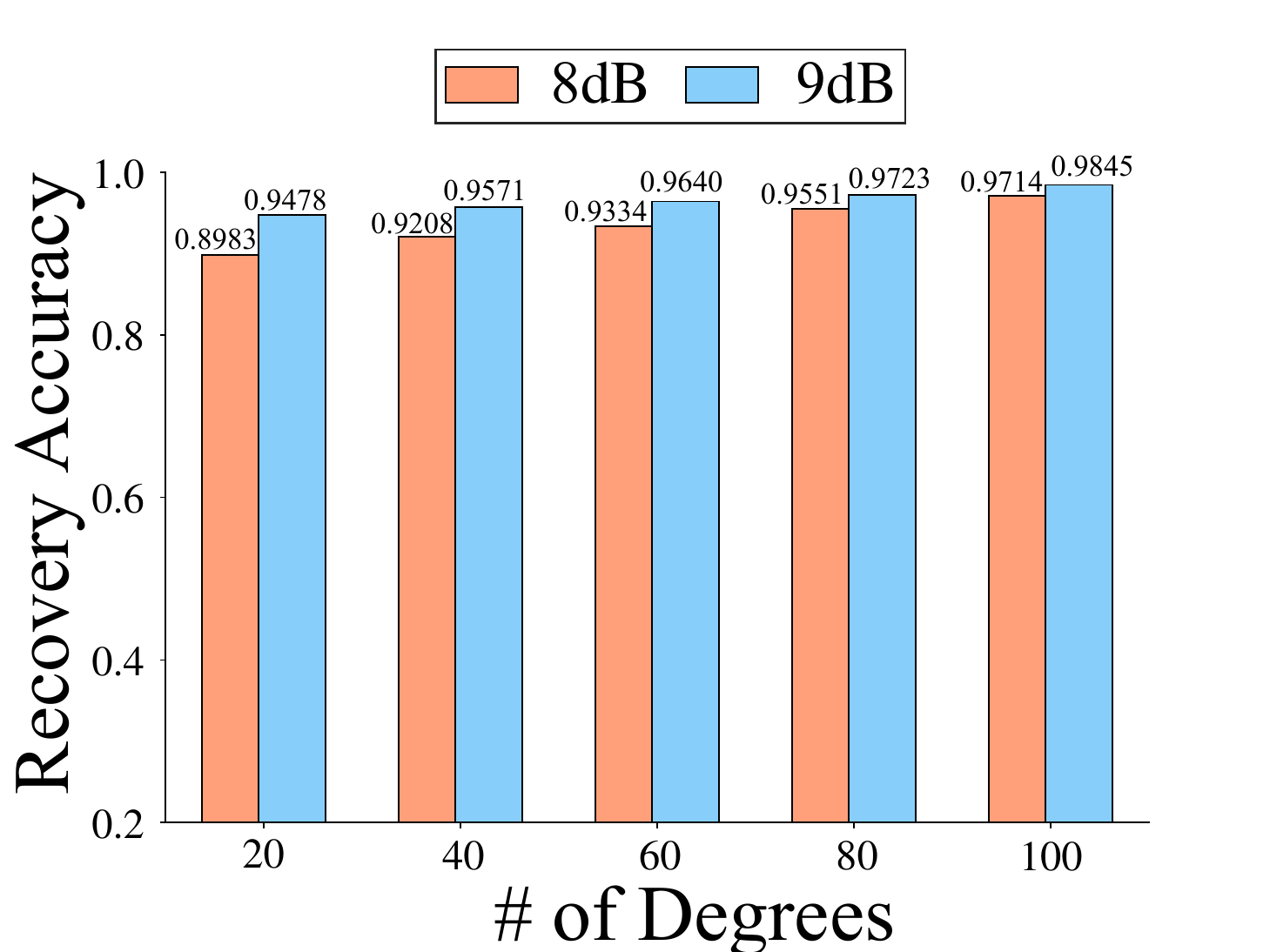}
    \captionsetup{labelformat=empty}
    \caption*{\footnotesize{(b)}}
    \label{Fig_ACCvsNumDegree_highSNR}
    \end{minipage}
\caption{\footnotesize{Accuracy of symbol recovery when recovering semantic symbols with different degrees under SNRs at (a) 2dB and 3dB, and (b) 8dB and 9dB.}}
    \label{Fig_ACCvsNumDegree}
\end{figure}
 \begin{figure}[htbp]
    \centering
    \begin{minipage}{.45\linewidth}
    \centering
    \includegraphics[width=\textwidth]{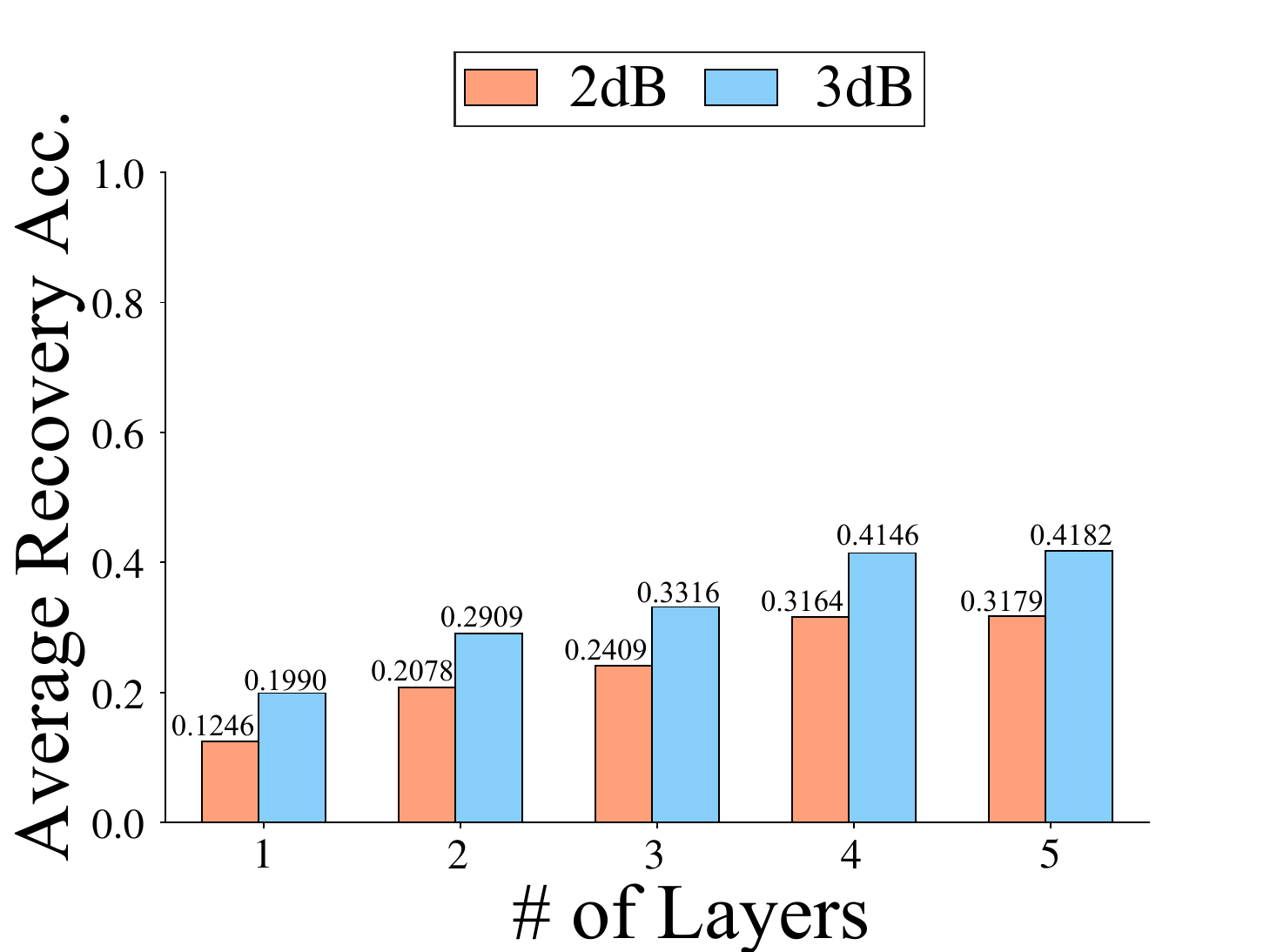}
    \vspace{-0.1in}
    \captionsetup{labelformat=empty}
    \vspace{-0.1in}
    \caption*{\footnotesize{(a)}}
    \end{minipage}
    \begin{minipage}{.45\linewidth}
    \centering
    \includegraphics[width=\textwidth]{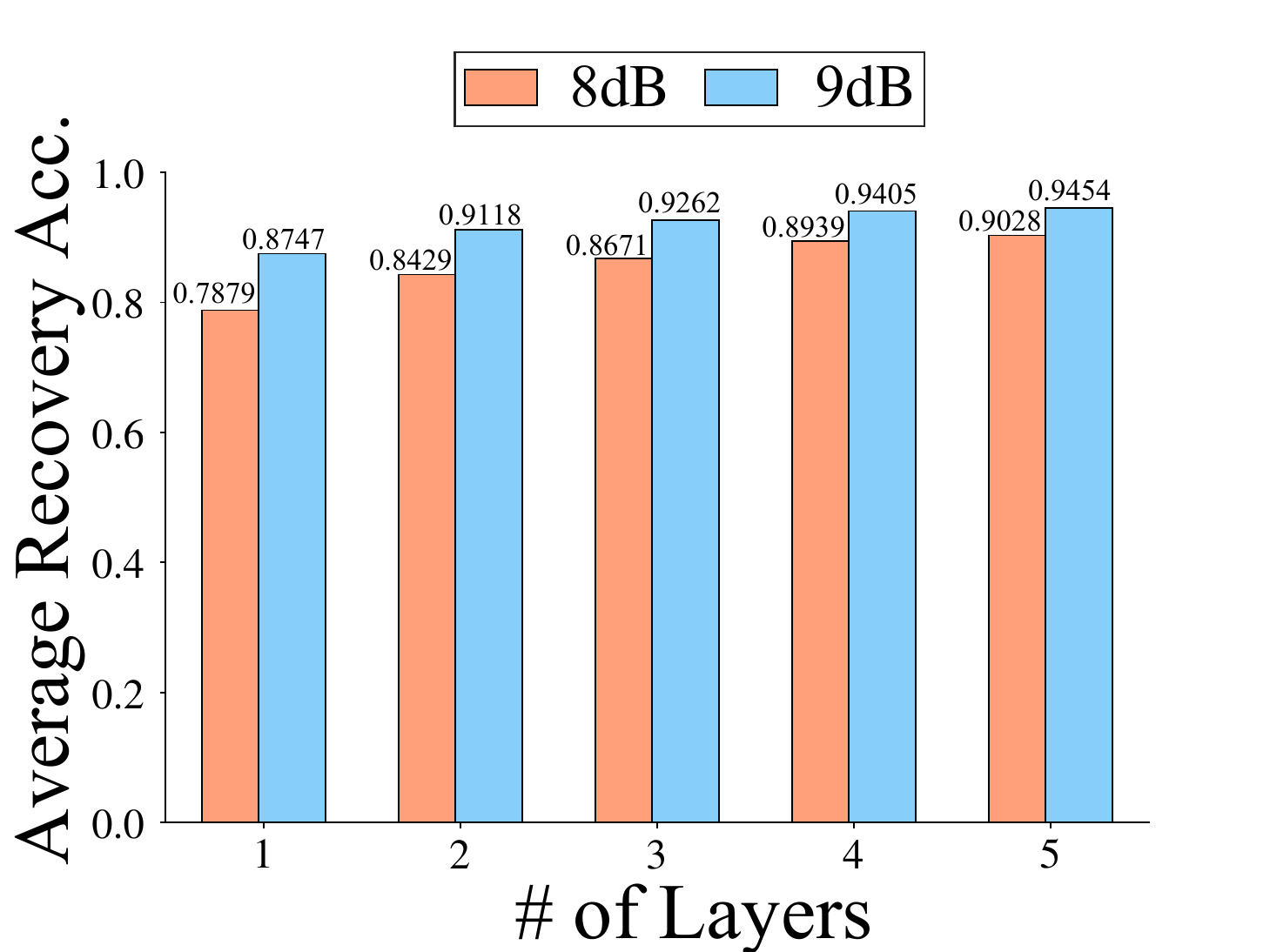}
    \vspace{-0.1in}
    \captionsetup{labelformat=empty}
    \vspace{-0.1in}
    \caption*{\footnotesize{(b)}}
    \end{minipage}
    \vspace{-0.1in}
\caption{\footnotesize{Average symbol recovery accuracy when the semantic entities have been divided into different layers of abstraction.}} 
\label{Fig_ACCvsNumLayers}
\end{figure}

\subsection{Numerical Results}

\subsubsection{Performance of Semantic Encoder/Decoder}
We first evaluate the performance of our proposed semantic encoding and decoding solutions for a single pair of source and destination users when communicating via noisy channels. 
In fact, once the semantic interpreter being trained, it can be used not only to infer the implicit semantics but also recover some missing semantics corrupted during the physical channel transmission. We therefore evaluate the semantic symbol error rate when the semantic interpreter can be used by the destination user to recover the corrupted semantic symbols (entities) that are associated with different abstraction layers sent over physical channel with additive Gaussian noises under different signal-to-noise ratio (SNR) in Fig. \ref{Fig_SymbolicErrorRate}. We simulate the semantics of messages using FB15K-237 dataset. We can observe that the symbol error performance of semantic symbols (entities) in different layers of abstractions is generally different. In particular, the higher-layer symbols generally have better error correction performance than that of the lower-layer symbols. For example, when the received SNR is 4dB, compared to the no semantic encoding/decoding solutions, our proposed semantic interpreter offers 17.3 dB, 17.9 dB, and 25.8 dB improvements on the symbol error rate when the entities are associated with low layer, mid layer, and high layer of abstractions when using the hard decoding as shown in Fig. \ref{Fig_SymbolicErrorRate}(a). This is because, as observed in our previous discussion, high layer semantic symbols generally have higher degrees and are densely connected with each other, 
compared to the low layer symbols. Therefore, when being corrupted in the noisy channel transmission, the high layer symbols have a higher chance to be recovered by other directly connected entities and relations.
In Fig. \ref{Fig_SymbolicErrorRate}(b), we also compare the symbol error rate when the corrupted semantic symbols can be recovered with soft decoding. 
We can observe that, in this case, 
our proposed semantic interpreter achieves 16.3 dB, 16.9 dB, and 24.7 dB improvements on the semantic symbol error rate when the entities are associated with low layer, mid layer, and high layer of abstractions, respectively, compared to the no semantic reasoning-based symbol recovery solution. 

In Fig. \ref{Fig_ACCvsNumDegree}, we consider semantics of messages generated from 13,621 entities in FB15K-237 dataset with degrees that are less than or equal to 100, and then evaluate the semantic symbol recovery accuracy when the semantic interpreter has been applied to recover semantic entities with different degrees. We can observe that our proposed semantic interpreter offers a better semantic recovery performance for semantic symbols with higher degree numbers, especially when the SNR is low. When the SNR increases into a large value, e.g., 8dB or 9dB, the impact of the semantic symbol's degree on the recovery accuracy becomes limited, i.e., when degrees increase from 20 to 100, only 8\% and 4\% improvements are achieved in semantic symbol recovery accuracy under SNRs 8dB and 9dB, respectively.  

\subsubsection{Performance of Multi-layer Representation}
In this paper, we propose multi-layer representation of semantics that is suitable to be implemented into a multi-tier cloud/edge networking architecture.
In our proposed representation, semantic entities have been categorized into multiple layers based on the abstraction levels of their represented meanings. In some human knowledge databases, the difference in the meaning abstractions can be reflected by the degree of the semantic entities, e.g., entities with higher abstraction levels may have higher degrees. To evaluate the impact of the multi-layered representations on the semantic communication performance, we once again focus on the semantic entities in FB15K-237 dataset with degrees that are less than or equal to 100 and investigate the cases when the semantic entities can be divided into different numbers of layers according to their degrees, e.g., when the number of layers is 1, all the entities belong to the same layer, when the number of layers is 2, the semantic entities with degrees 0-50 and 51-100 will be categorized into different layers, similarly, when the numbers of layers are 3, (4 or 5), the entities with degrees 0-30, 31-60, and 61-100 (0-30, 31-60, 61-80, and 81-100 or 0-20, 21-40, 41-60, 61-80, and 81-100) will be categorized into 3 (4, or 5) different layers. In Fig. \ref{Fig_ACCvsNumLayers}, we compare the average semantic symbol recovery accuracy when the semantics can be represented by the above different layered structure. We can observe that the layered representation of semantics can further improve the error correction performance of semantic communications, especially when the SNR is low. This is because, as suggested by Fig. \ref{Fig_ACCvsNumDegree}, the higher layered entities (with higher degrees) tend to enjoy better robustness against channel corruption, and therefore by dividing the semantic entities into multiple layers, the improved recovery accuracy of the higher layered entities will benefit the recovery performance of their descendant entities.


\begin{figure}[htbp]
\centering
    \begin{minipage}[t]{.45\linewidth}
    \centering
    \includegraphics[width=\textwidth]{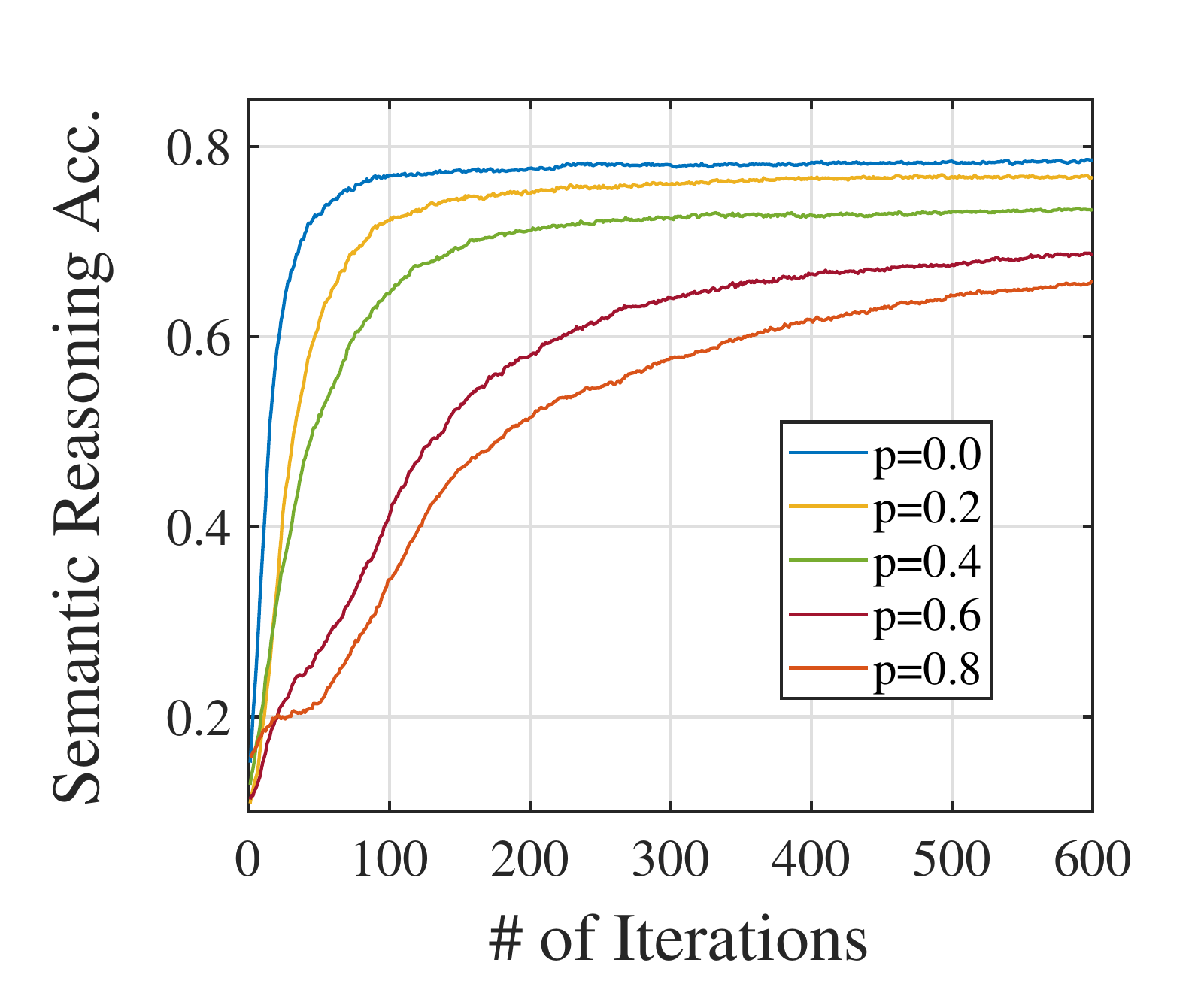}
    \vspace{-0.2in}
    \subcaption{}
    \label{Fig_ValidationAccuracy_SamplingMechanism_cora}
    \end{minipage}
    \begin{minipage}[t]{0.45\linewidth}
    \centering
    \includegraphics[width=\textwidth]{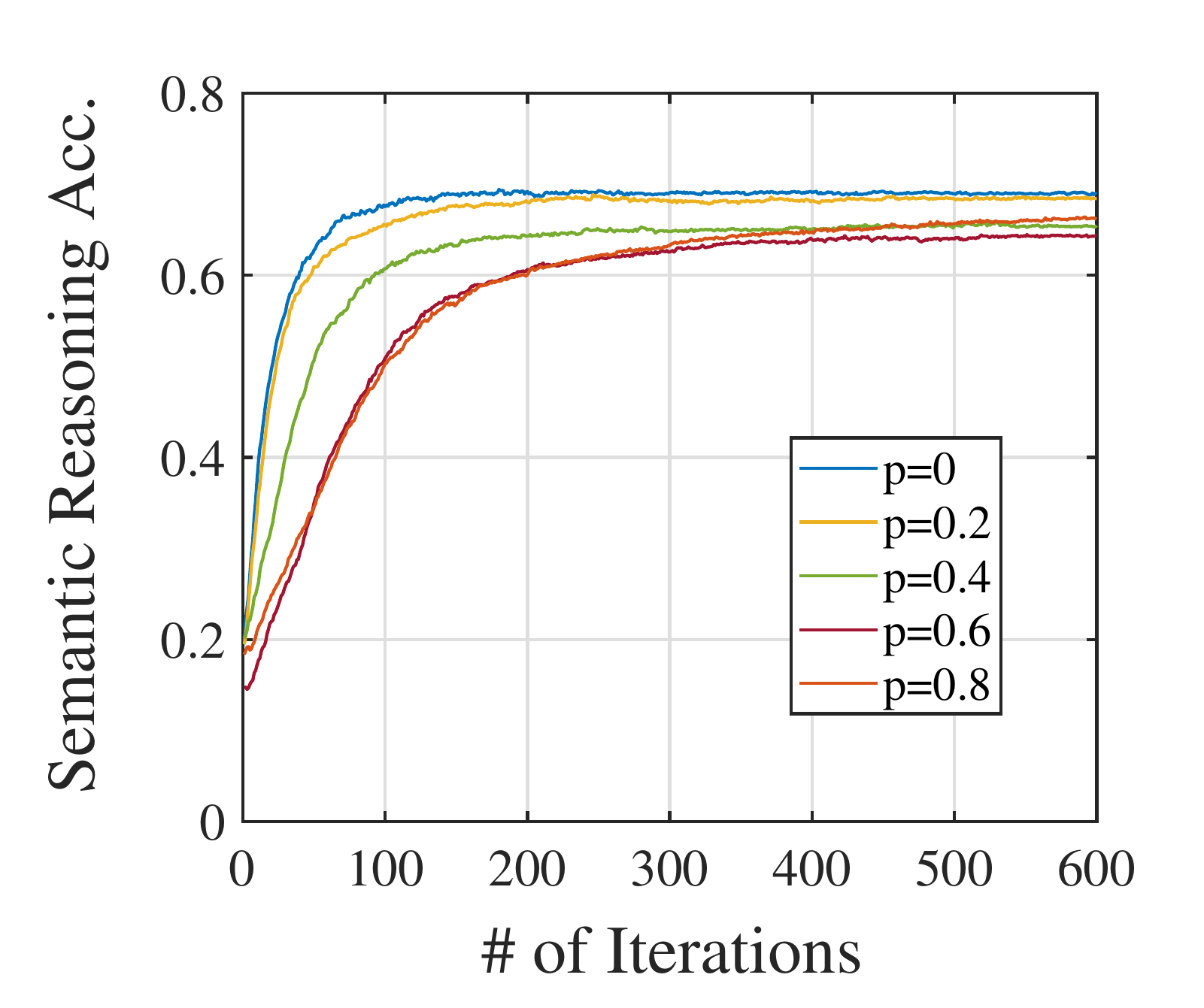}
    \vspace{-0.2in}
    \subcaption{}
    \label{Fig_ValidationAccuracy_SamplingMechanism_citeseer}
    \end{minipage}
    \vspace{-0.1in}
\caption{\footnotesize{Semantic reasoning accuracy of collaborative reasoning under different training iterations with different i.i.d. datasets based on (a) Cora and (b) Citeseer datasets.}} 
\label{Fig_ValidationAccuracy_SamplingMechanism}
\end{figure}

\begin{figure}[htbp]
    \centering
    \begin{minipage}[t]{.45\linewidth}
    \centering
    \includegraphics[width=\textwidth]{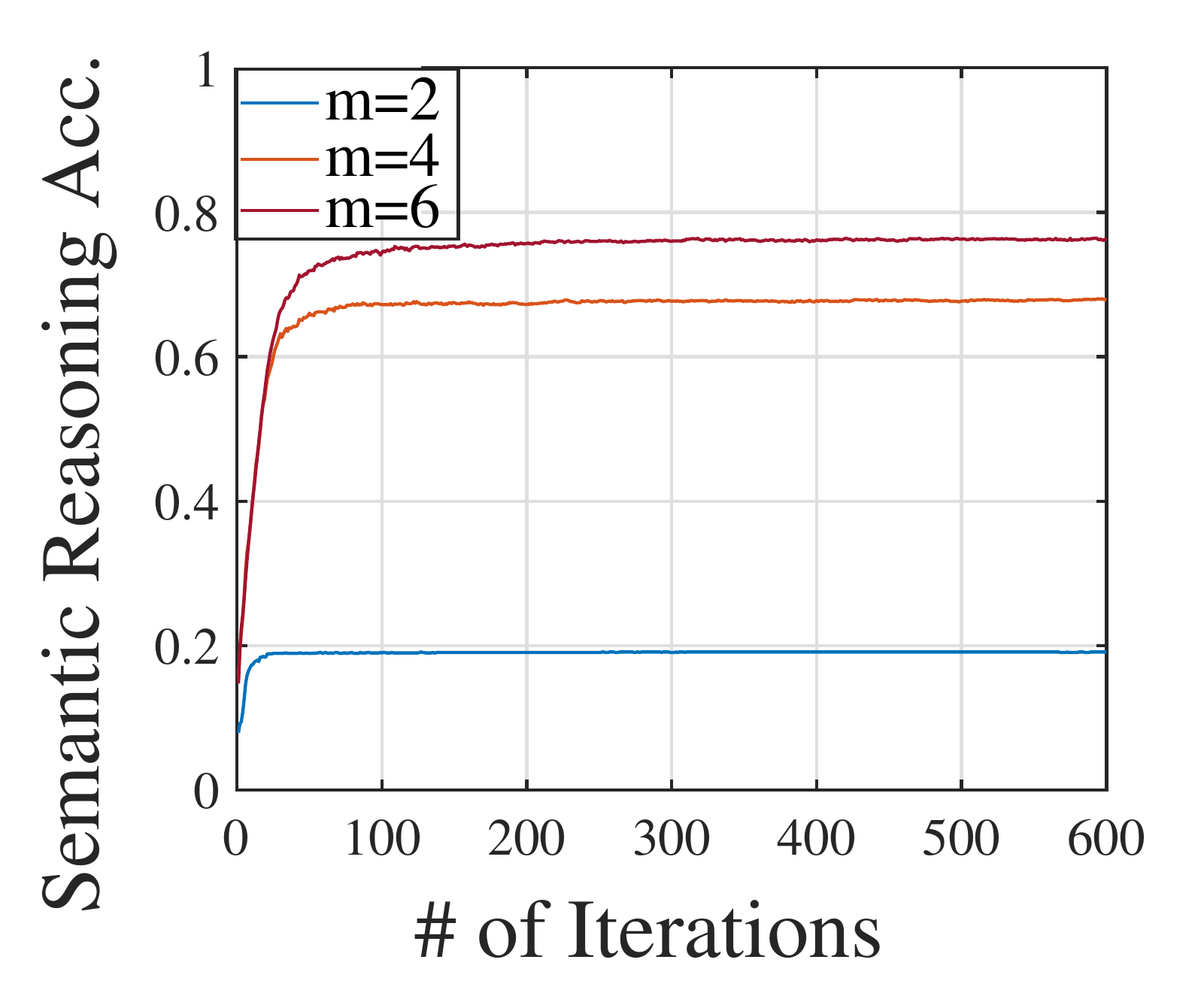}
    \label{Fig_ValidationAccuracy_UserNumber_SamplingMechanism_p0}
    \vspace{-0.2in}
    \captionsetup{labelformat=empty}
    \caption*{\footnotesize{(a) Cora, $p=0$}} 
    \end{minipage}
    \begin{minipage}[t]{.45\linewidth}
    \centering
    \includegraphics[width=\textwidth]{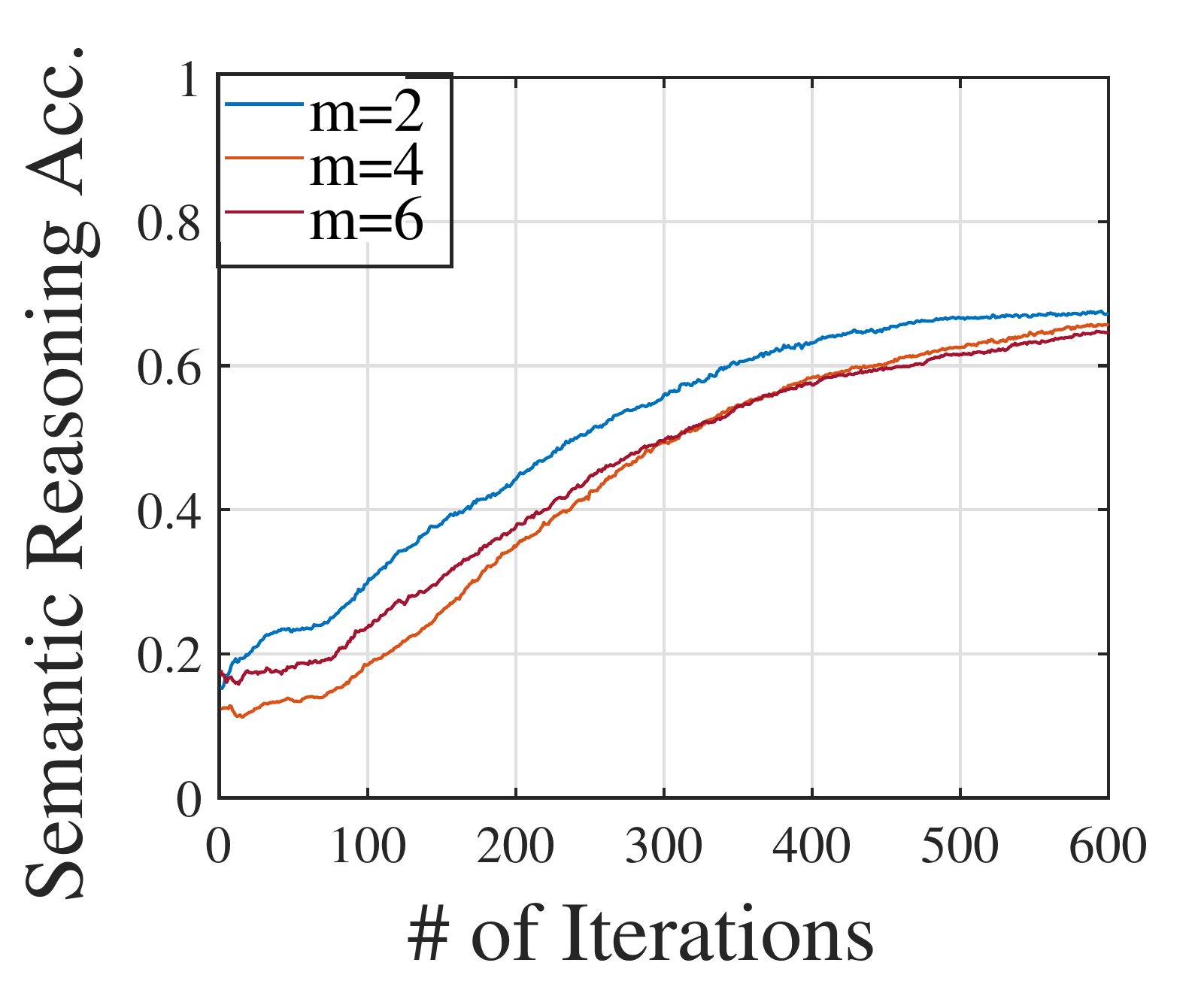}
    \vspace{-0.2in}
    \captionsetup{labelformat=empty}
    \caption*{\footnotesize{(b) Cora, $p=1$}} 
    \label{Fig_ValidationAccuracy_UserNumber_SamplingMechanism_p1.0}
    \vspace{-0.2in}
    \end{minipage}
    \begin{minipage}[t]{.45\linewidth}
    \centering
    \includegraphics[width=\textwidth]{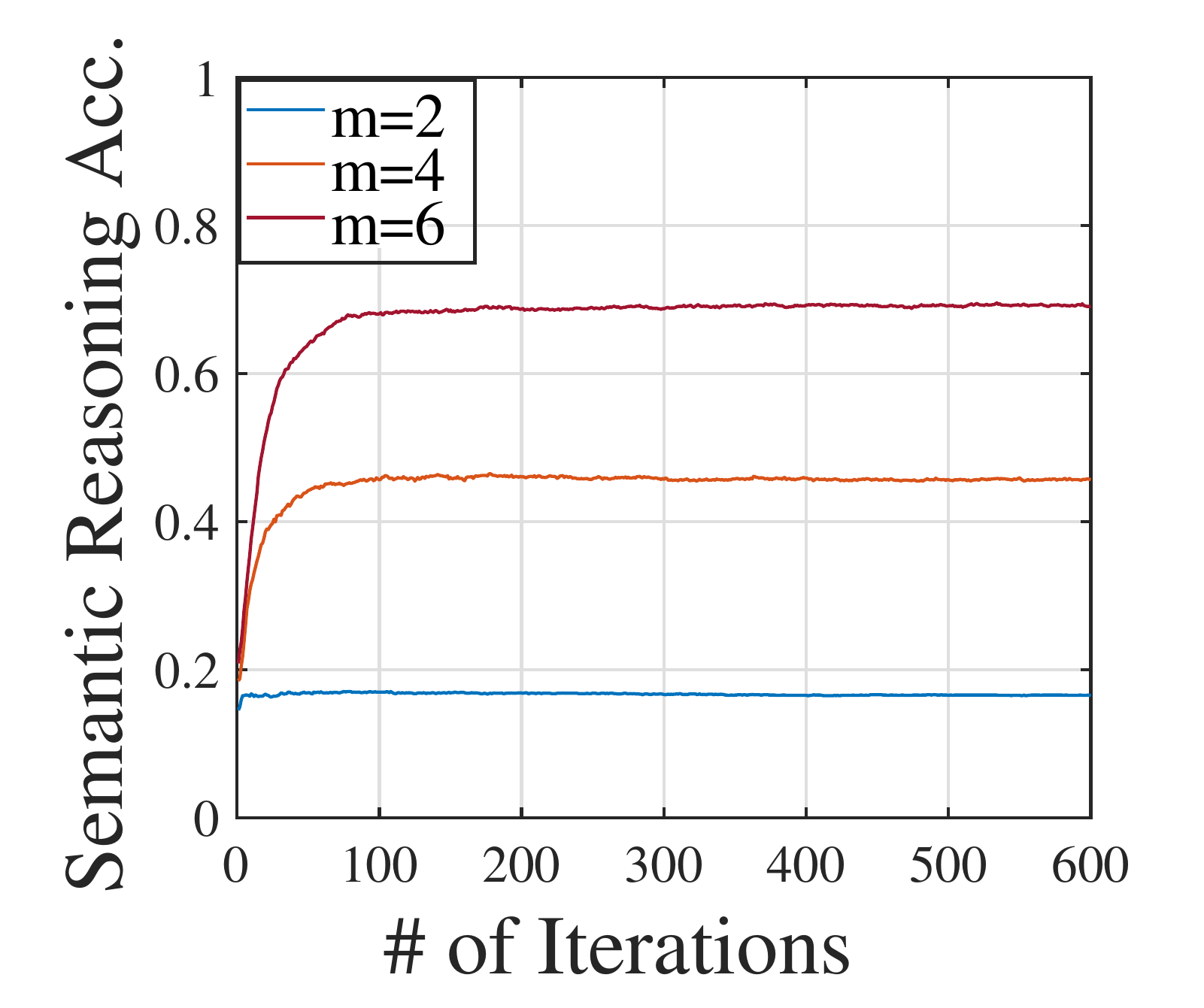}
    \captionsetup{labelformat=empty}
    \vspace{-0.2in}
    \caption*{\footnotesize{(c) Citeseer, $p=0$}} 
    \label{Fig_ValidationAccuracy_UserNumber_SamplingMechanism_p0_citeseer}
    \end{minipage}
    \begin{minipage}[t]{.45\linewidth}
    \centering
    \includegraphics[width=\textwidth]{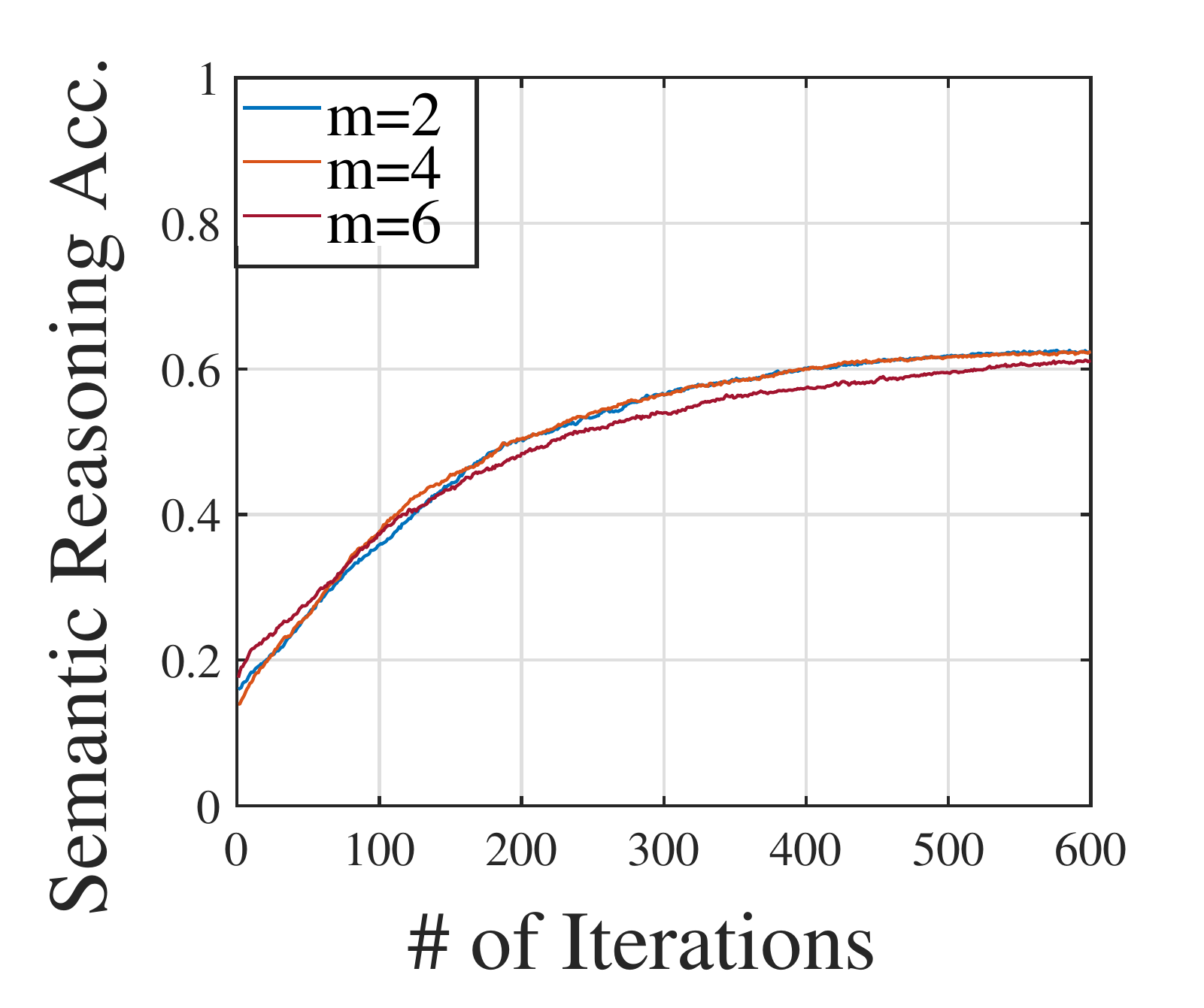}
    \captionsetup{labelformat=empty}
    \vspace{-0.2in}
    \caption*{\footnotesize{(d) Citeseer, $p=1$}} 
    \label{Fig_ValidationAccuracy_UserNumber_SamplingMechanism_p1.0_citeseer}
    \end{minipage}
\caption{\footnotesize{Semantic reasoning accuracy under different training iterations with different numbers of collaborative edge servers with different non-i.i.d. datasets.} }
\label{Fig_ValidationAccuracy_UserNumber_SamplingMechanism}
\end{figure}

\subsubsection{Performance of Collaborative Reasoning}
Let us now evaluate the performance of our proposed collaborative reasoning approach. 
As mentioned earlier, the heterogeneity  of decentralized datasets at different edge servers may directly affect the convergence of the  model training of the  collaborative reasoning. To simulate different levels of heterogeneity between datasets, we consider different distribution schemes of knowledge entities associated with 6 different subjects in two citation datasets Cora and Citeseer\cite{sen2008CoraCiteseer}. In particular, we divide the entities in the datasets of each edge server into two parts: one part consists of entities associated with a single subject that can be exclusively accessed by the edge server, the other part consists of entities that are uniformly random sampled from a combination of entities across 6 subjects. We introduce a random variable $p$ to denote the portion of the first part entities in the dataset of each edge server, entities 
associated with a single subject with exclusive accessibility. For example, $p=1$ means that all the entities in each edge server are uniformly randomly sampled from all 6 subjects of knowledge entities. $p=0.5$ means that a half of entities in the datasets of each edge server is associated with a single subject and the rest half of entities are randomly sampled from the combination of entities randomly sampled from 6 subjects. $p=0$ means that the dataset in each edge server only has entities associated with a single subject. In this way, the datasets in different edge serves can be considered as i.i.d. (or non-i.i.d.) when $p=1$ (or $p=0$). All other values of $p$ are associated with different non-i.i.d. degrees of datasets across different edge servers.

In Fig. \ref{Fig_ValidationAccuracy_SamplingMechanism}, we compare the convergence performance of collaborative reasoning based on datasets with different non-i.i.d. degree. We can observe that when the non-i.i.d. degree of datasets at different edge servers increases, the convergence rate of collaborative model training can be improved for both considered knowledge datasets. Also, the improvement of convergence rate caused by the increase of non-i.i.d. degree of datasets is higher in Cora than that in Citeseer. This is because the knowledge entities in Cora has relatively  higher degrees of relations in average compared to the entities in CiteSeer. In other words, the convergence performance of collaborative reasoning exhibits stronger relationship with the  non-i.i.d. degree of  decentralized datasets when the knowledge entities are more densely connected to each other. 

In Fig. \ref{Fig_ValidationAccuracy_UserNumber_SamplingMechanism}, we compare the convergence speed of collaborative reasoning under different numbers of collaborative edge servers. We can observe that when the datasets are non-i.i.d. ($p=0$), the convergence speed increases significantly when the number of edge servers increases from 2 to 6. However, if the dataset in each edge servers are uniformly sampled from all the 6 subjects of knowledge entities (i.i.d. datasets), the convergence improvement achieved by increasing the number of collaborative edge servers becomes limited.

\begin{figure}[htbp]
    \centering
    \begin{minipage}{.45\linewidth}
    \centering
    \centering
    \includegraphics[width=\textwidth]{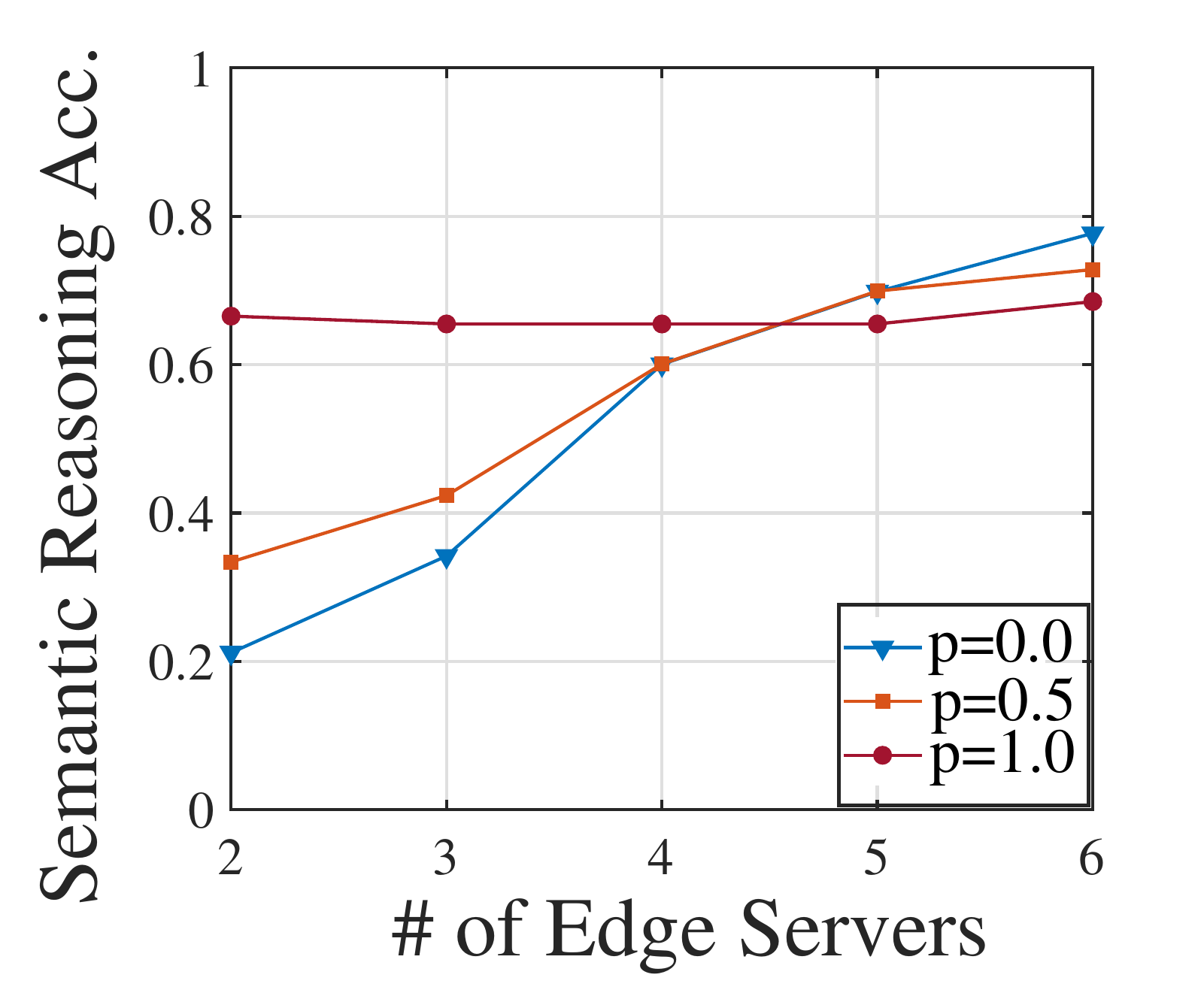}
    \vspace{-0.2in}
    \captionsetup{labelformat=empty}
    \vspace{-0.1in}
    \caption*{\footnotesize{(a)}}
    \label{Fig_TestAccuracy_UserNumber_cora}
    \end{minipage}
    \begin{minipage}{.45\linewidth}
    \centering
    \includegraphics[width=\textwidth]{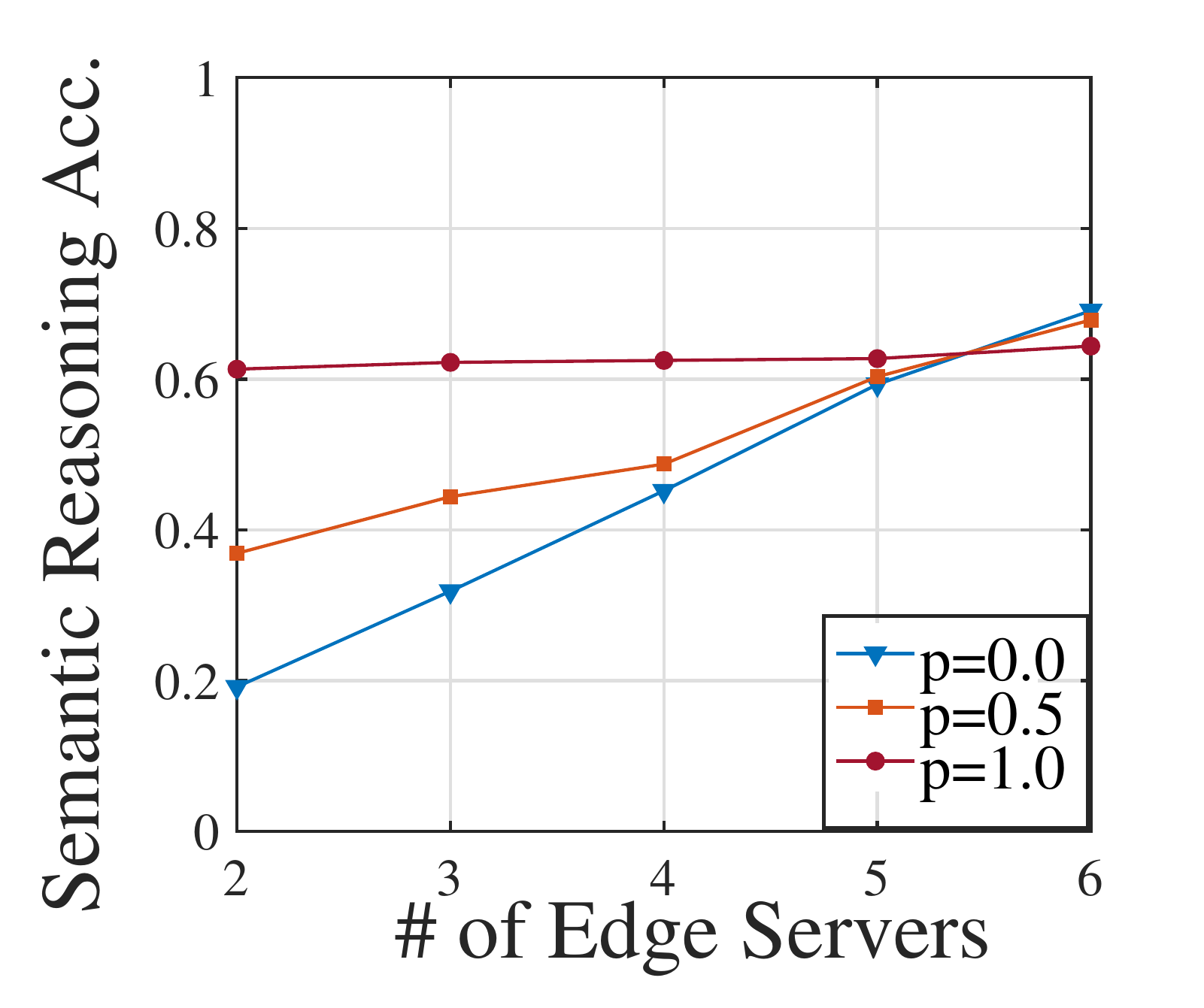}
    \vspace{-0.2in}
    \captionsetup{labelformat=empty}
    \vspace{-0.1in}
    \caption*{\footnotesize{(b)}}
    \label{Fig_TestAccuracy_UserNumber_citeseer}
    \end{minipage}
    \vspace{-0.1in}
\caption{\footnotesize{Semantic reasoning accuracy of model trained with different number of edge servers.}}
\label{Fig_TrainingLoss&TestAccuracy_UserNumber}
\end{figure}

In Fig. \ref{Fig_TrainingLoss&TestAccuracy_UserNumber}, we compare the converged semantic recovery accuracy after 600 local SGD iterations when the semantic reasoning model is trained with different numbers of collaborative edge servers. We can observe that the  accuracy of the trained model always increases with the numbers of collaborative edge servers. Also, when the number of edge servers increases, the increasing rate of the model accuracy trained with non-i.i.d datasets will be higher than that with i.i.d datasets. We can also observe that different knowledge datasets exhibit different increasing rates of trained model accuracy when the number of edge servers increases. In particular,  in Cora dataset, the accuracy of the trained model increases at a higher speed when the number of collaborative edge servers is small, e.g., increases from 2 to 4, and the increasing rate of the model accuracy becomes slower when the number of edge servers becomes larger than or equal to  5. In the Citeseer dataset however, the model accuracy increases almost linearly with the number of edge servers.

\vspace{-0.2in}
\section{Conclusion}
\label{Section_Conclusion}

This paper has proposed a novel collaborative reasoning-based implicit semantic communication architecture that allow users and edge servers to jointly learn to imitate the implicit semantic reasoning behavior of the source users to support efficient semantic encoding, decoding, and interpretation. In particular, a new multi-layer representation of semantics taking into consideration of both a hierarchy of implicit semantics across different abstraction levels as well as personally preferred semantic reasoning mechanism of individual user has been proposed. We have then proposed a multi-tier collaborative reasoning-based semantic communication architecture in which the users rely on CDC and edge servers for encoding, decoding, and interpreting semantic meaning involving knowledge concepts stored at CDC and/or edge servers. An imitation-based reasoning mechanism learning solution has been developed for the source users to assist  CDC and edge servers to train semantic reasoning mechanism models that can imitate the true inference mechanism of the source users. A federated GCN-based collaborative reasoning solution has proposed to allow same-tier edge servers to joint construct a shared model to generate implicit semantics from the observed explicit semantics. Finally, extensive experiments have been conducted to evaluate the performance of our proposed solution. Numerical results confirm that our proposed solution can achieve over 82.9\% accuracy level for implicit semantic recovery at the destination user, even communicated in noisy faded physical channel condition.

\section*{Acknowledgment}
Y. Xiao and G. Shi were supported in part by the major key project of Peng Cheng Laboratory (No. PCL2021A12). Y. Xiao was also supported in part by the National Natural Science Foundation of China under grant 62071193 and the Key R \& D Program of Hubei Province of China under grants 2021EHB015 and 2020BAA002. G. Shi was also supported in part by the National Natural Science Foundation of China under grants 61871304, 61976169. 

\begin{appendices}
\section{Proof of Theorem \ref{Theorem_Convexity}}\label{Proof_Convexity}
To prove 
the optimization problem in (\ref{eq_MaxLagCauEntropyProblem}) is a strongly convex problem, let us first present the following lemmas that are necessary for our proof of Theorem \ref{Theorem_Convexity}.
\subsection{Key Lemmas}
\begin{lemma}\label{Lemma_AdditiveConvexity}
If a function $F(x) = H(x) + \Gamma(x)$, where $H$ is strongly convex and $\Gamma$ is convex, then $F$ is strongly convex.
\end{lemma}

\begin{lemma}\label{Lemma_EntropyStronglyConvex}
$\bar{H}(c)=\sum_{s,a}{c(a, s)\log(c(a, s)/\sum_{a'} c(a', s))}$ is strongly convex given the bounded occupancy measure, i.e., there exist two positive constants $M$ and $\xi$, such that $c(a, s) \leq M$ and $c(a, s)/\sum_{a'} c(a', s) \leq 1-\xi$.
\end{lemma}

\subsection{Proof of Lemma \ref{Lemma_AdditiveConvexity}}
\begin{proof}
Suppose $m>0$ and $H$ is $m$-strongly convex, then by definition, for all valid $x$ and $y$, we have:
\begin{eqnarray}\label{eq_Lemma_AdditiveConvextiy_StronglyConvex}
\bar{H}\left(y\right) \geq \bar{H}\left(x\right)+ \nabla \bar{H}(x)^{T}\left(y-x\right) +   \frac{m}{2} \ |y-x\|^{2},
\end{eqnarray}
where $\nabla \bar{H}(x)$ is the gradient at $x$ if $\bar{H}$ is differentiable, and it can be replace by sub-gradient when the function is non-differentiable.
And if $\Gamma$ is convex, by definition, we can write:
\begin{eqnarray}\label{eq_Lemma_AdditiveConvextiy_Convex}
\Gamma\left(y\right) \geq \Gamma\left(x\right) + \nabla \Gamma^{T}\left(x\right)\left(y-x\right).
\end{eqnarray}
By adding eqn.(\ref{eq_Lemma_AdditiveConvextiy_StronglyConvex}) and eqn.(\ref{eq_Lemma_AdditiveConvextiy_Convex}), we have:
\begin{eqnarray}
\bar{H}\left(y\right) + \Gamma\left(y\right)&& \geq \;\;\; \bar{H}\left(x\right) + \Gamma\left(x\right) \nonumber \\
&& + \left(\nabla\bar{H}(x) + \nabla\Gamma\left(x\right)\right)^{T}\left(y-x\right) \nonumber \\
&& + \frac{m}{2} \ |y-x\|^{2},
\end{eqnarray}
which can be written as:
\begin{eqnarray}
F\left(y\right)  \geq F\left(x\right) + \nabla F^{T}\left(x\right)\left(y-x\right)+ \frac{m}{2} \ |y-x\|^{2}.
\nonumber
\end{eqnarray}
Hence, $F$ is $m$-strongly convex, which completes the proof.
\end{proof}

\subsection{Proof of Lemma \ref{Lemma_EntropyStronglyConvex}}
\begin{proof}
Let $\bar{h}(c) = c(a, s)\log(c(a, s)/\sum_{a'} c(a', s))$, by taking the first derivative with respect to $c$, we have
\begin{eqnarray}
\nabla \bar{h}(c) = \log c\left(a, s\right) + 1 - \log \sum_{a'} c\left(a', s\right) - \frac{c\left(a, s\right)}{\sum_{a'} c\left(a', s\right)}.
\nonumber
\end{eqnarray}
Then, we take the second derivative, and rearrange it as follows:

\begin{eqnarray}\label{eq_Lemma_EntropyStronglyConvex_SecondDerivative}
\begin{aligned}
\nabla^2 \bar{h}(c)
&& = \;\;\; \frac{1}{c\left(a, s\right)} - \frac{2}{c\left(a, s\right)\sum_{a'} c\left(a', s\right)} \\
&& + \frac{c\left(a, s\right)}{\left(\sum_{a'} c\left(a',s\right)\right)^2}  \\
&& = \frac{1}{c\left(a, s\right)}\left(1-\frac{c\left(a, s\right)}{\sum_{a'} c\left(a', s\right)}\right)^2
\end{aligned}
\end{eqnarray}

We have
\begin{eqnarray}
\label{eq_EntropyStronglyConvex_BoundedPolicy}
\frac{1}{c\left(a, s\right)} \geq \frac{1}{M} \;\;\;\mbox{ and }\;\;\; \left(1-\frac{c\left(a, s\right)}{\sum_{a'} c\left(a', s\right)}\right)^2 \geq \xi^2.
\end{eqnarray}

By combining (\ref{eq_Lemma_EntropyStronglyConvex_SecondDerivative})-(\ref{eq_EntropyStronglyConvex_BoundedPolicy}), we have
\begin{eqnarray}
\nabla^2 \bar{h}(c) \geq \frac{\xi^2}{M},
\end{eqnarray}
which proves that $\bar{h}$ is $\bar{\mu}$-strongly convex and $\bar{\mu}=\xi^2/M$.
Note that $\bar{H}$ is sum of $\bar{h}$. We can therefore conclude that $\bar{H}$ is $\mu$-strongly convex where $\mu$ is a positive constant. 
\end{proof}

\subsection{Proof of Theorem \ref{Theorem_Convexity}}
\begin{proof}
By definition, we have
\begin{eqnarray}
\begin{aligned}
H(\pi) &=\mathbb{E}_{\pi}[-\log \pi(a, s)]=-\sum_{s, a} c_{\pi}(a, s) \log \pi(a, s) \\
&=-\sum_{s, a} c_{\pi}(a, s) \log \frac{c_{\pi}(a, s)}{\sum_{a^{\prime}} c_{\pi}\left(a^{\prime}, s\right)}=-\bar{H}\left(c_{\pi}\right).
\end{aligned}
\end{eqnarray}
From Lemma \ref{Lemma_EntropyStronglyConvex}, $-H(\pi)$ is $\mu$-strongly convex. In addition, $\Gamma(\pi_E, \pi_D)$ is in the form of cross-entropy or TransE that has been proved to be convex. From Lemma \ref{Lemma_AdditiveConvexity}, we can conclude that the sum of $H(\pi)$ and $\Gamma(\pi_E, \pi_D)$ is also $\mu$-strongly convex, which completes the proof.
\end{proof}
\vspace{-0.2in}
\section{Proof of Theorem \ref{Theorem_GAN}}\label{Proof_GAN}
Due to the limit of space, we present a brief description of proof of Theorem \ref{Theorem_GAN}.  
We adopt the idea similar to a modified version of generative adversarial networks (GAN) in which the interpreter (generator) and the evaluator (discriminator) can be trained in a competitive fashion\cite{wang2020mathematical}. In particular,  
as shown in Theorem \ref{Theorem_Convexity}, the objective function described in problem (\ref{eq_ProblemInterpreter}) is strongly convex, which means that we can find the optimal $\varpi_{\phi}^{*}$ by taking the first derivative of the objective function and equating it to zero. Then we can use SGD-based method to obtain the optimal $\mathbf{\pi}_{D}$ that minimizes the objective function in problem (\ref{eq_SemanticEvaluator}) taking value at $\varpi_{\phi}^{*}$. In other words, the KL divergence between the distributions of expert paths and reasoning paths generated by the interpreter can minimized, which concludes our proof. 

\vspace{-0.2in}
\section{Proof of Theorem \ref{Theorem_ConvergeFL}}\label{Proof_ConvergeFL}
\subsection{Key Lemmas}
We first present several key lemmas that are necessary to the proofs of Theorem \ref{Theorem_ConvergeFL}. Let $\mathbf{g}_{t}=\sum_{k=1}^{K} \gamma_{k} \nabla F_{k}\left(\mathbf{w}_{t}^{k}\right)$ be the aggregated gradients.

\begin{lemma}\label{Lemma_ExpectedPara}
Let Assumptions \ref{assumption_continous}-\ref{assumption_BoundedGradient} hold. If we choose the learning rate $\eta_t \leq \frac{1}{4L}$, we have
\begin{eqnarray}
\begin{aligned}
\mathbb{E}\|\mathbf{\bar{w}}_{t+1}-\mathbf{w}^{*}\|^{2}
&& \leq\left(1-\mu \eta_{t}\right)\mathbb{E}\|\mathbf{\bar{w}}_{t}-\mathbf{w}^{*}\|^{2}\\
&& +4{\eta_t}^2\left(E-1\right)^2{\sigma_L}^2 \\
&& +4 \eta_{t}^{2} L \rho +\eta_{t}{ }^{2} \mathbb{E}\|\mathbf{g}_{t}\|^{2} \nonumber
\end{aligned}
\end{eqnarray}
\end{lemma}

\begin{lemma}\label{Lemma_GradientBound}
Assume \cite[Theorem 5.4]{yao2022fedgcn} and Assumption \ref{assumption_BoundedGradient} hold, we have
\begin{eqnarray}
\mathbb{E}\|\mathbf{g}_{t}\|^2 \leq \frac{2L_{p}}{N}\mathcal{D} + 2{\sigma_L}^2 \nonumber
\end{eqnarray}
where $\mathcal{D}=\left\|K \mathbf{X}_{k}^{T} \mathbf{A}_{k}^{T} \mathbf{A}_{k}^{T} \mathbf{A}_{k} \mathbf{A}_{k} \mathbf{X}_{k}-\mathbf{X}^{T} \mathbf{A}^{T} \mathbf{A}^{T} \mathbf{A} \mathbf{A} \mathbf{X}\right\|^2$.
\end{lemma}

\subsection{Proof of Lemma \ref{Lemma_ExpectedPara}}
\begin{proof}
Notice that $\mathbf{\bar{w}}_{t+1}=\mathbf{\bar{w}}_{t}-{\eta}_t\mathbf{g}_t$, then we have

\begin{eqnarray}\label{Lemma_ExpectedPara_eqn1}
\begin{aligned}
& \left\|\mathbf{\bar{w}}_{t+1}-\mathbf{w}^{*}\right\|^{2} \\
&= \left\|\mathbf{\bar{w}}_{t}-\eta_{t} \mathbf{g}_{t}-\mathbf{w}^{*}\right\|^{2} \\
& = \left\|\mathbf{\bar{w}}_{t}-\mathbf{w}^{*}\right\|^{2} +\underbrace{2 \eta_{t}\left\langle\mathbf{\bar{w}}_{t}-\mathbf{w}^{*}, \mathbf{g}_{t}\right\rangle}_{A_{1}}+\left\|\eta_{t} \mathbf{g}_{t}\right\|^{2}.
\end{aligned}
\end{eqnarray}

Next, we focus on bounding $A_1$ that can be split into two parts:

\begin{eqnarray}\label{Lemma_ExpectedPara_eqn2}
\begin{aligned}
A_{1} &=-2 \eta_{t}\left\langle\mathbf{\bar{w}}_{t}-\mathbf{w}^{*}, \mathbf{g}_{t}\right\rangle\\
& =-2 \eta_{t} \sum_{k=1}^{K} \gamma_{k}\left\langle\mathbf{\bar{w}}_{t}-\mathbf{w}^{*}, \nabla F_{k}\left(\mathbf{w}_{t}^{k}\right)\right\rangle \\
& =-2 \eta_{t} \sum_{k=1}^{K} \gamma_{k}\langle\underbrace{\left\langle\mathbf{\bar{w}}_{t}-\mathbf{w}_{t}^{k}, \nabla F_{k}\left(\mathbf{w}_{t}^{k}\right)\right\rangle}_{B_{1}}\\
&\;\;\;\;-2 \eta_{t} \sum_{k=1}^{K} \gamma_{k}\langle\underbrace{\left.\mathbf{w}_{t}^{k}-\mathbf{w}^{*}, \nabla F_{k}\left(\mathbf{w}_{t}^{k}\right)\right\rangle}_{B_{2}}.
\end{aligned}
\end{eqnarray}
By applying the Cauchy-Schwarz inequality and AM-GM, $B_1$ can be bounded as:

\begin{eqnarray}\label{Lemma_ExpectedPara_eqn3}
\begin{aligned}
-2 B_{1} & =-2\left\langle\mathbf{\bar{w}}_{t}-\mathbf{w}_{t}^{k}, \nabla F_{k}\left(\mathbf{w}_{t}^{k}\right)\right\rangle \\
& \leq \frac{1}{\eta_{t}}\left\|\mathbf{\bar{w}}_{t}-\mathbf{w}_{t}^{k}\right\|^{2}+\eta_{t}\left\|\nabla F_{k}\left(\mathbf{w}_{t}^{k}\right)\right\|^{2}.
\end{aligned}
\end{eqnarray}
From the $\mu$-strong convexity of $F_k(\cdot)$, $B_2$ can be bounded as:

\begin{eqnarray}\label{Lemma_ExpectedPara_eqn4}
\begin{aligned}
-B_{2}& =\left\langle \mathbf{w}_{t}^{k}-\mathbf{w}^{*}, \nabla F_{k}\left(\mathbf{w}_{t}^{k}\right)\right\rangle\\
& \leq-\left(F_{k}\left(\mathbf{w}_{t}^{k}\right)-F_{k}\left(\mathbf{w}^{*}\right)\right)-\frac{\mu}{2}\left\|\mathbf{w}_{t}^{k}-\mathbf{w}^{*}\right\|.
\end{aligned}
\end{eqnarray}
By the $L$-smoothness of $F_k(\cdot)$, we have

\begin{eqnarray}\label{Lemma_ExpectedPara_eqn5}
\left\|\nabla F_{k}\left(\mathbf{w}_{t}^{k}\right)\right\|^{2} \leq 2 L\left(F_{k}\left(\mathbf{w}_{t}^{k}\right)-F_{k}^{*}\right).
\end{eqnarray}
By combining eqn.(\ref{Lemma_ExpectedPara_eqn1})-(\ref{Lemma_ExpectedPara_eqn5}), we have

\begin{eqnarray}\label{Lemma_ExpectedPara_eqn6}
\begin{aligned}
\left\|\mathbf{\bar{w}}_{t+1}-\mathbf{w}^{*}\right\|^{2} \leq &\left(1-\mu \eta_{t}\right)\left\|\mathbf{\bar{w}}_{t}-\mathbf{w}^{*}\right\|^{2} \\
& +\sum_{k=1}^{K} \gamma_{k}\left\|\mathbf{\bar{w}}_{t}-\mathbf{w}_{t}^{k}\right\|^{2} \\
& +A_2+\left\|\eta_{t} \mathbf{g}_{t}\right\|^{2}.
\end{aligned}
\end{eqnarray}
where
\begin{equation*}
\begin{aligned}
A_2 = & 2 L \eta_{t}^{2} \sum_{k=1}^{K} \gamma_{k}\left(F_{k}\left(\mathbf{w}_{t}^{k}\right)-F_{k}^{*}\right)\\
& -2 \eta_{t} \sum_{k=1}^{K} \gamma_{k}\left(F_{k}\left(\mathbf{w}_{t}^{k}\right)-F_{k}\left(\mathbf{w}^{*}\right)\right)
\end{aligned}
\end{equation*}
According to \cite[Lemma 1]{li2019convergence}, $A_2$ can be further bounded as:

\begin{eqnarray}
\begin{aligned}
A_2 \leq 4 L \eta_{t}^{2} \rho+\sum_{k=1}^{K} \gamma_{k}\left\|\mathbf{\bar{w}}_{t}-\mathbf{w}_{t}^{k}\right\|^{2}. \nonumber
\end{aligned}
\end{eqnarray}
Substituting $A_2$ into eqn.(\ref{Lemma_ExpectedPara_eqn6}), taking the expectations on both sides, and assume \cite[Lemma 3]{li2019convergence} holds, we then have

\begin{eqnarray}
\begin{aligned}
\mathbb{E}\left\|\mathbf{\bar{w}}_{t+1}-\mathbf{w}^{*}\right\|^{2} \leq & \left(1-\mu \eta_{t}\right) \mathbb{E}\left\|\mathbf{\bar{w}}_{t}-\mathbf{w}^{*}\right\|^{2}\\
& +4{\eta_t}^2\left(E-1\right)^2{\sigma_L}^2+4 L \eta_{t}^{2} \rho\\
&+\mathbb{E}\left\|\eta_{t} \mathbf{g}_{t}\right\|^{2}.
\end{aligned}
\end{eqnarray}

This concludes the proof. 
\end{proof}

\subsection{Proof of Lemma \ref{Lemma_GradientBound}}
\begin{proof}
We first split $\mathbb{E}\|\mathbf{g}_t\|^2$ as follows:

\begin{eqnarray}\label{Lemma_GradientBound_eqn1}
\begin{aligned}
\mathbb{E}\|{\mathbf{g}_t}\|^2 &= \mathbb{E}\left\|\mathbf{g}_{t}-\nabla F_{k}\left(\mathbf{w}_{t}^{k}\right)+\nabla F_{k}\left(\mathbf{w}_{t}^{k}\right)\right\|^{2}\\
&\leq 2 \mathbb{E}\left\|g_{t}-\nabla F_{k}\left(w_{t}^{k}\right)\right\|^{2}+2 \sigma_{L}^{2},
\end{aligned}
\end{eqnarray}
where the last inequality is based on Assumption \ref{assumption_BoundedGradient} and the fact that, for random variables $z_1,...,z_m$, we have
\begin{equation*}
\mathbb{E}\left[\left\|z_{1}+\cdots+z_{m}\right\|^{2}\right] \leq m \mathbb{E}\left[\left\|z_{1}\right\|^{2}+\cdots+\left\|z_{m}\right\|^{2}\right].
\end{equation*}
Following the same line as \cite[Theorem 5.4]{yao2022fedgcn}, we can rewrite (\ref{Lemma_GradientBound_eqn1}) as

\begin{eqnarray}\label{Lemma_GradientBound_eqn2}
\mathbb{E}\left\|\mathbf{g}_{t}\right\|^{2} \leq \frac{2 L_{p}}{N}\mathcal{D}+2 \sigma_{L}^{2}.
\end{eqnarray}
This concludes our  proof. 
\end{proof}

\subsection{Proof of Theorem \ref{Theorem_ConvergeFL}}
\begin{proof}
By combining Lemma \ref{Lemma_ExpectedPara} and Lemma \ref{Lemma_GradientBound}, we can have
\begin{eqnarray}
\mathbb{E}\|\mathbf{\bar{w}}_{t+1}-\mathbf{w}^*\|^2 \leq \left(1-\mu \eta_{t}\right) \mathbb{E}\left\|\mathbf{\bar{w}}_{t}-\mathbf{w}^{*}\right\|^{2}+{\eta_t}^2 Y,
\end{eqnarray}
where
\begin{eqnarray}
Y = 2 ((E-1)^{2}+1) \sigma_{L}^{2}+4  L \rho + \frac{2 L_{p}}{N}\mathcal{D}.
\end{eqnarray}
Following method adopted in \cite[Theorem 1]{li2019convergence} and substituting $\Omega=4\left(1+2(E-1)^{2}\right) \sigma_{L}^{2}+4 L \rho+\frac{\mu^{2} \zeta}{4}\left\|\mathbf{w}_{1}-\mathbf{w}^{*}\right\|^{2}$, we can have

\begin{eqnarray}
\mathbb{E}\left[F\left(\mathbf{\bar{w}}_{T}\right)\right]-F^{*} \leq \frac{2 \kappa}{\zeta+T-1}\left(\frac{\Omega}{\mu}+\frac{2 L_{p}}{\mu N}\mathcal{D}\right),
\end{eqnarray}
where $\eta_{t}=\frac{2}{\mu} \frac{1}{\zeta+t}$, $\zeta=\max \{8 \kappa, E\}$, and $\kappa=\frac{L}{\mu}$. This concludes the proof.
\end{proof}
\end{appendices}

\bibliographystyle{IEEEtran}
\bibliography{bibtex}

\begin{IEEEbiography}[{\includegraphics[width=1.1in,height=1.3in,clip,keepaspectratio]{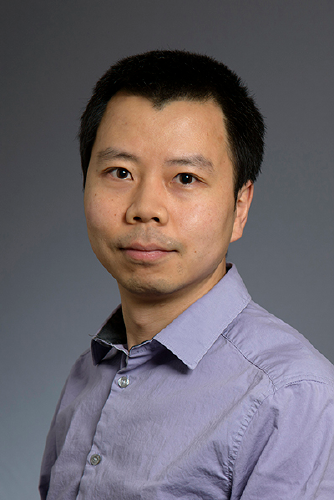}}]{Yong Xiao} (Senior Member, IEEE) received his B.S. degree in electrical engineering from China University of Geosciences, Wuhan, China in 2002, M.Sc. degree in telecommunication from Hong Kong University of Science and Technology in 2006, and his Ph. D degree in electrical and electronic engineering from Nanyang Technological University, Singapore in 2012. He is now a professor in the School of Electronic Information and Communications at the Huazhong University of Science and Technology (HUST), Wuhan, China. He is also with Peng Cheng Laboratory, Shenzhen, China and Pazhou Laboratory (Huangpu), Guangzhou, China. He is the associate group leader of the network intelligence group of IMT-2030 (6G promoting group) and the vice director of 5G Verticals Innovation Laboratory at HUST. Before he joins HUST, he was a  research assistant professor in the Department of Electrical and Computer Engineering at the University of Arizona where he was also the center manager of the Broadband Wireless Access and Applications Center (BWAC), an NSF Industry/University Cooperative Research Center (I/UCRC) led by the University of Arizona. His research interests include machine learning, game theory, distributed optimization, and their applications in cloud/fog/mobile edge computing, green communication systems, wireless communication networks, and Internet-of-Things (IoT).
\end{IEEEbiography}

\begin{IEEEbiography}[{\includegraphics[width=1.1in,height=1.3in,clip,keepaspectratio]{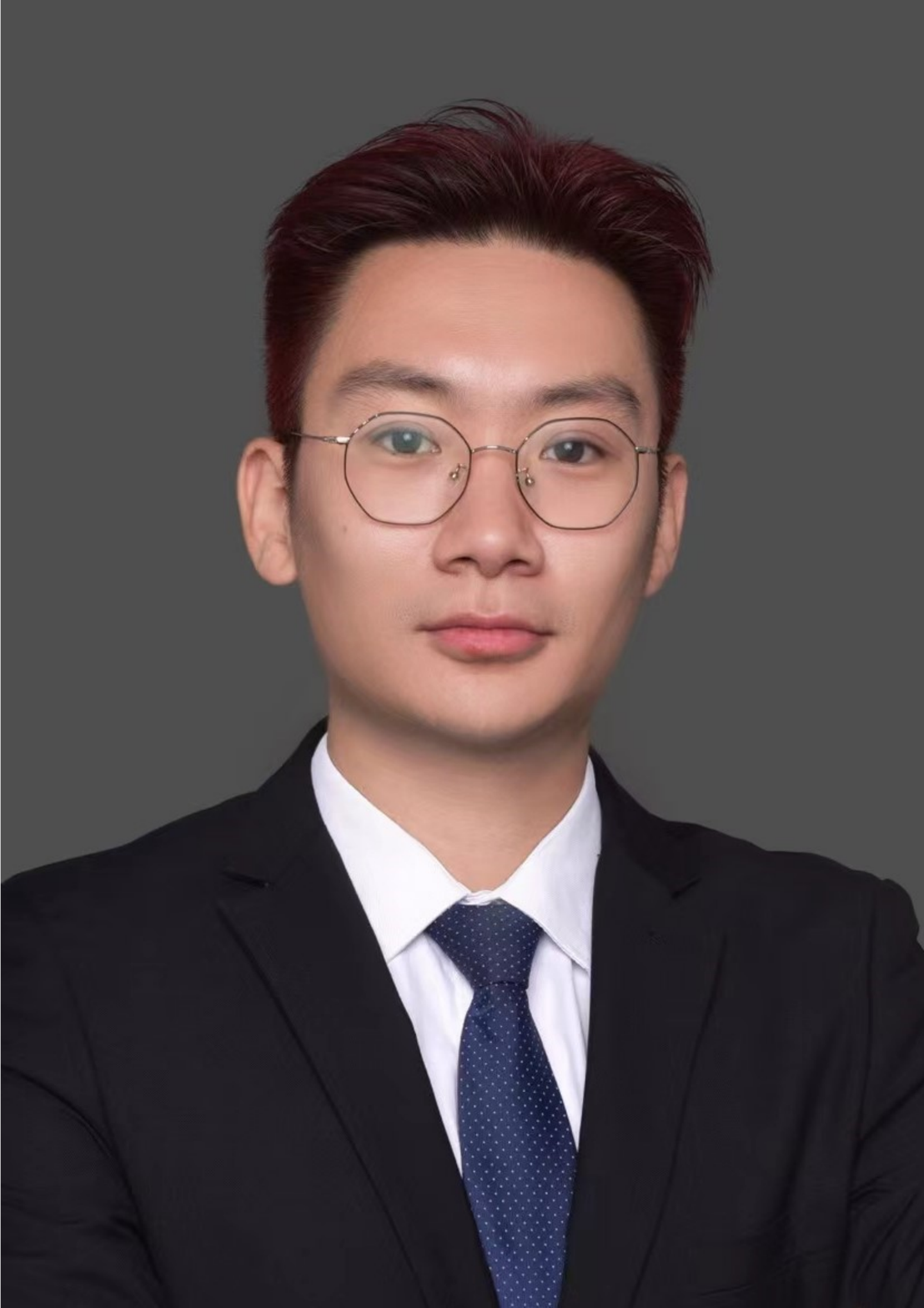}}]{Zijian Sun} (Student Member, IEEE) received his B.S. degree in information engineering from Huazhong University of Science and Technology, Wuhan, China in 2014, and M.S. degree in electrical engineering from the University of Melbourne, Australia in 2017. He is currently pursuing his PhD in the school of electronic information and communications at the Huazhong University of Science and Technology, Wuhan, China. His research interest includes network AI and next generation communication technology.
\end{IEEEbiography}

\begin{IEEEbiography}[{\includegraphics[width=1.1in,height=1.3in,clip,keepaspectratio]{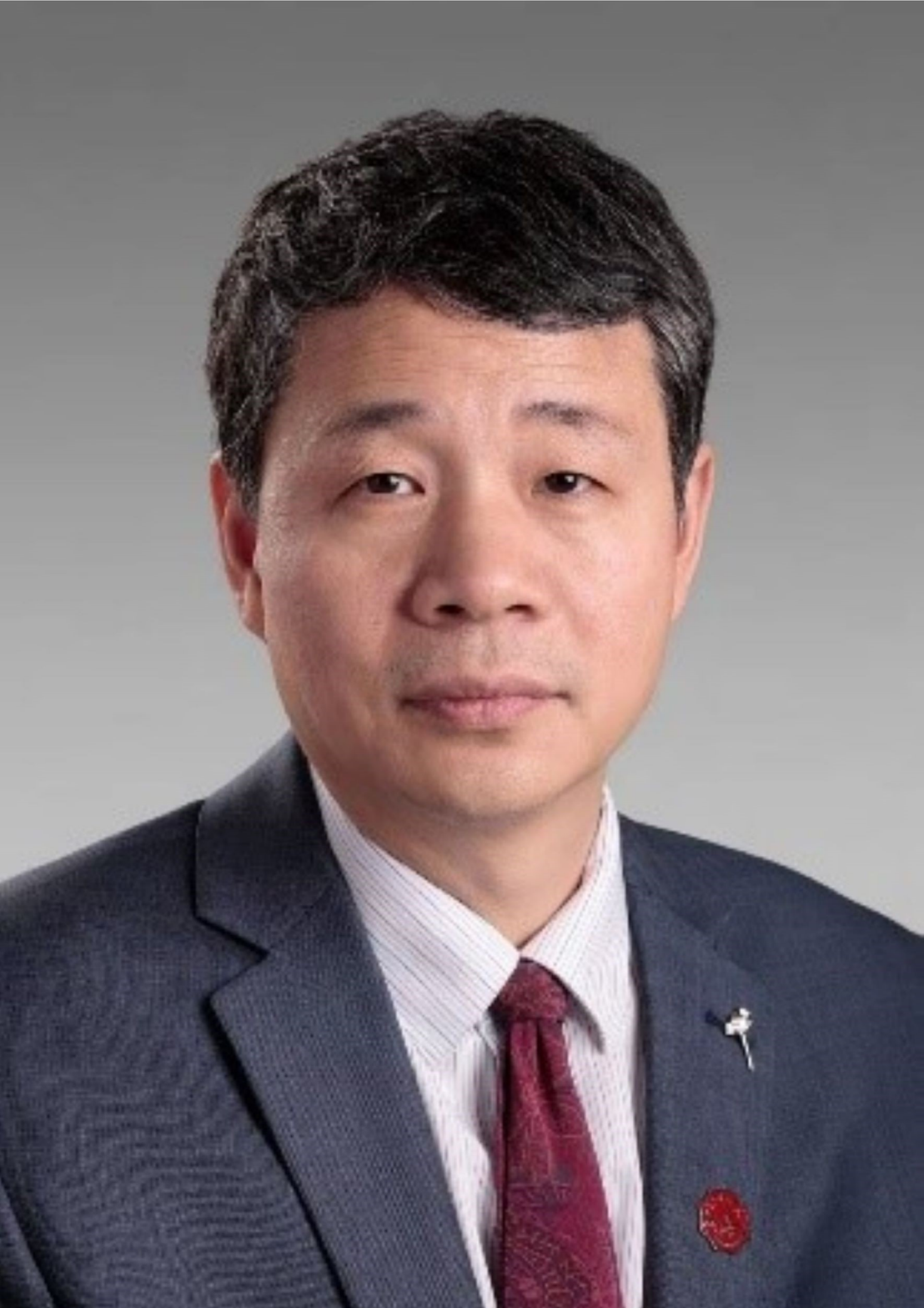}}]{Guangming Shi} (Fellow, IEEE) received the M.S. degree in computer control, and the Ph.D. degree in electronic information technology from Xidian University, Xi’an, China, in 1988, and 2002, respectively. He was the vice president of Xidian University from 2018 to 2022. Currently, he is  the Vice Dean of Peng Cheng Laboratory and a Professor with the School of Artificial Intelligence, Xidian University. He is an IEEE Fellow, the chair of IEEE CASS Xi’an Chapter, senior member of ACM and CCF, Fellow of Chinese Institute of Electronics, and Fellow of IET. He was awarded Cheung Kong scholar Chair Professor by the ministry of education in 2012. He won the second prize of the National Natural Science Award in 2017. His research interests include Artificial Intelligence, Semantic Communications, and Human-Computer Interaction.
\end{IEEEbiography}

\begin{IEEEbiography}[{\includegraphics[width=1.1in,height=1.3in,clip,keepaspectratio]{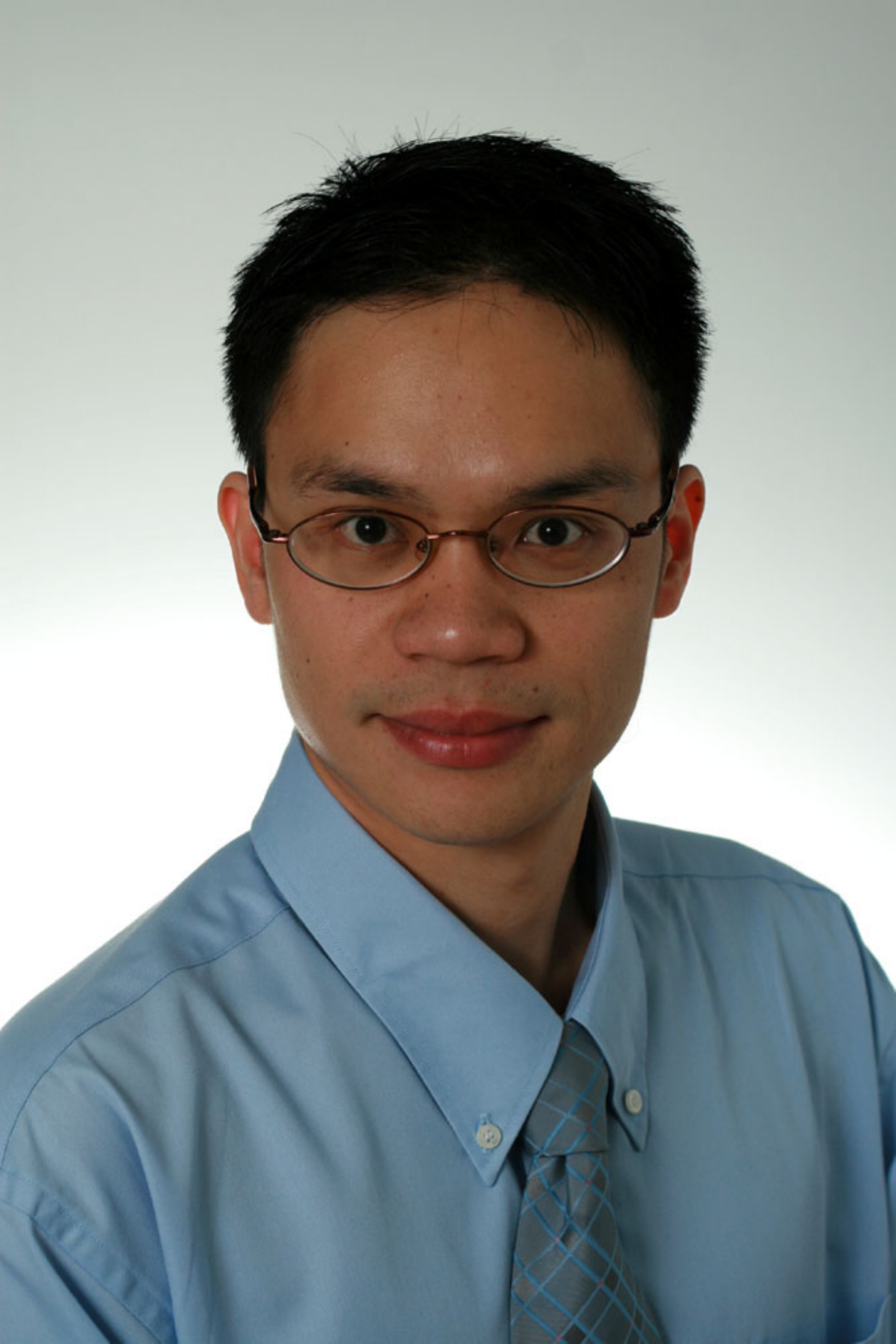}}]{Dusit Niyato} (Fellow, IEEE) is a professor in the School of Computer Science and Engineering, at Nanyang Technological University, Singapore. He received B.Eng. from King Mongkuts Institute of Technology Ladkrabang (KMITL), Thailand in 1999 and Ph.D. in Electrical and Computer Engineering from the University of Manitoba, Canada in 2008. His research interests are in the areas of Internet of Things (IoT), machine learning, and incentive mechanism design.
\end{IEEEbiography}

\end{document}